%% file: neurips_2026.tex
\documentclass{article}

\PassOptionsToPackage{numbers,compress}{natbib}

\usepackage[preprint]{neurips_2026}

\usepackage[utf8]{inputenc} 
\usepackage[T1]{fontenc}    
\usepackage{hyperref}       
\usepackage{url}            
\usepackage{booktabs}       
\usepackage{amsfonts}       
\usepackage{amsmath}        
\usepackage{nicefrac}       
\usepackage{microtype}      
\usepackage[table]{xcolor} 
\usepackage{graphicx}       
\usepackage{multirow}       
\usepackage{pifont}         

\usepackage{array}
\usepackage{xurl}
\usepackage{enumitem}
\usepackage{float}
\usepackage{placeins}
\usepackage{tabularx}
\usepackage{makecell}
\newcolumntype{Y}{>{\centering\arraybackslash}X}

\newcolumntype{C}[1]{>{\centering\arraybackslash}m{#1}}
\newcommand{\cmark}{\ding{51}}
\newcommand{\xmark}{\ding{55}}

\title{AgroVG: A Large-Scale Multi-Source Benchmark for Agricultural Visual Grounding}

\author{%
\textbf{Haocheng Li}$^{1}$,
\textbf{Juepeng Zheng}$^{2,6,\dagger}$,
\textbf{Zenghao Yang}$^{3}$,
\textbf{Kaiqi Du}$^{1}$, \\
\textbf{Guilong Xiao}$^{1}$,
\textbf{Gengmeng Pu}$^{1}$,
\textbf{Haohuan Fu}$^{4,6}$,
\textbf{Jianxi Huang}$^{1,5,\dagger}$ \\
\\[-1mm]
\multicolumn{1}{c}{\normalfont\small
$^{1}$China Agricultural University \quad
$^{2}$Sun Yat-sen University \quad
$^{3}$Tianjin University} \\
\multicolumn{1}{c}{\normalfont\small
$^{4}$Tsinghua University \quad
$^{5}$Southwest Jiaotong University} \\
\multicolumn{1}{c}{\normalfont\small
$^{6}$National Supercomputing Center in Shenzhen}
}

\begin{document}
\maketitle

\begin{abstract}
Visual grounding, the task of localizing objects described by natural-language expressions, is a foundational capability for agricultural AI systems, enabling applications such as selective weeding, disease monitoring, and targeted harvesting. Reliable evaluation of agricultural visual grounding remains challenging because agricultural targets are often small, repetitive, occluded, or irregularly shaped, and instructions may refer to one, many, or no objects in an image. Evaluating this capability therefore requires jointly testing localization accuracy, target-set completeness, and existence-aware abstention. To address these challenges, we introduce \textbf{AgroVG}, a multi-source benchmark that formulates agricultural grounding as generalized set prediction: given an image and a referring expression, a model must return all matching target instances or abstain when no target is present. AgroVG contains 10{,}071 annotation-grounded image-query pairs from ten source datasets across six target families: crop/weed, fruit, wheat head, pest, plant disease, and tree canopy. It supports bounding-box grounding (T1) across all six families and instance-mask grounding (T2) on sources with reliable instance-level pixel annotations, with queries covering single-target, multi-target, and target-absent regimes. AgroVG further provides task-specific protocols for box-set matching and query-level mask coverage. Zero-shot evaluation of 26 model configurations spanning closed-source MLLMs, open-source VLMs, and specialized grounding systems reveals persistent gaps: the best multi-target Set-$F_1$ reaches only 0.35, and the best positive-query mask success rate at IoU@0.75 remains below 0.17. Data and code are available at \url{https://anonymous.4open.science/r/AgroVG-5172/}.
\end{abstract}

\vspace{-0.6em}

\begingroup
\setlength{\abovecaptionskip}{1pt}
\setlength{\belowcaptionskip}{-6pt}

\begin{figure}[H]
  \centering
  \includegraphics[
    width=0.98\textwidth,
    height=0.21\textheight,
    keepaspectratio,
    trim={0.15cm 0.10cm 0.15cm 0.10cm},
    clip
  ]{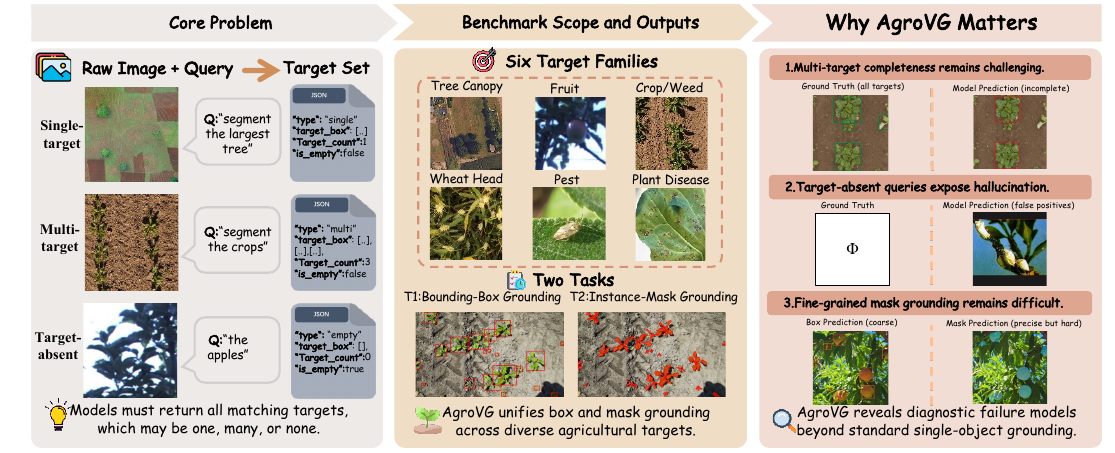}
  \caption{
  \textbf{AgroVG at a glance.}
  AgroVG casts agricultural grounding as generalized set prediction: a query over an agricultural image may require one target, multiple targets, or no output. The benchmark spans six target families and two output protocols, bounding boxes (T1) and instance masks (T2), to diagnose multi-target completeness, target-absent abstention, and mask grounding.
  \label{fig:teaser}}
\end{figure}

\endgroup


\section{Introduction}
\label{sec:intro}

As agriculture moves toward automated intervention and precision monitoring, AI systems must increasingly ground human instructions in complex crop environments. Imagine an agricultural robot instructed to remove ``all mature weeds among soybean plants,'' or a disease-monitoring system asked to localize ``every yellowing leaf in the upper canopy.'' Executing such instructions requires more than image-level recognition; the system must determine \emph{which} objects satisfy a natural-language description, \emph{where} they are located, and \emph{whether} such targets exist at all. This capability, commonly studied as visual grounding through referring expression comprehension and segmentation~\cite{kazemzadeh_referitgame_2014,mao_generation_2016}, is becoming an important enabling capability for agricultural AI systems, including instruction-following robots, precision monitoring tools, and broader vision-language-action pipelines~\cite{vougioukas_agricultural_2019,kim_openvla_2024}.

Despite rapid progress in visual grounding, existing benchmarks do not provide this evaluation setting for agricultural imagery. General-domain benchmarks have evolved from single-target referring expressions on everyday objects~\cite{yu_modeling_2016,mao_generation_2016} to generalized settings with multi-target and target-absent queries~\cite{he_grec_2023,liu_gres_2023}, while remote-sensing benchmarks evaluate language-conditioned localization in top-down aerial imagery~\cite{zhan_rsvg_2023,yuan_rrsis_2024,ding_vrsbench_2024}. These regimes differ substantially from close-range field, orchard, and canopy scenes. In agriculture, recent multimodal benchmarks evaluate recognition, visual question answering, captioning, and domain-specific reasoning~\cite{zhou_agribench_2024,shinoda_agrobench_2025,gauba_agmmu_2025,li_can_2025}, but they generally do not test whether a model can localize exact target regions, retrieve all matching instances, or abstain when no such target exists. The closest related agricultural grounding benchmark, gRef-CW~\cite{haghighat_multilabel_2026}, supports box-level generalized grounding on CropAndWeed imagery~\cite{steininger_cropandweed_2023}, but is restricted to a single source dataset, a single crop--weed sub-domain, and provides no instance-mask grounding evaluation. A unified benchmark that spans agricultural target families and jointly evaluates bounding-box and instance-mask grounding remains missing, as summarized in Table~\ref{tab:benchmark_comparison}.

Building a benchmark is non-trivial because agricultural grounding departs from the regime that drives existing benchmarks. Three properties make it distinct. \emph{(i) Dense, repetitive instances:} a single frame may contain dozens of visually similar wheat heads, clustered fruits, or interleaved crop and weed plants~\cite{david_global_2021,hani_minneapple_2020,steininger_cropandweed_2023}, so a referring expression cannot be assumed to map to a unique target. \emph{(ii) Small, occluded, and irregularly shaped targets:} pests, disease lesions, and individual leaves are often small or partially hidden, and many such targets have irregular, non-convex boundaries~\cite{wu_ip102_2019,wei_largescale_2026}, making bounding boxes insufficient for faithful pixel-level evaluation and motivating instance-mask grounding. \emph{(iii) Target-absent expressions:} realistic field instructions may refer to objects that do not appear in the observed image, requiring abstention rather than hallucinated predictions. Together, these properties motivate formulating agricultural grounding as a generalized set-prediction problem~\cite{liu_gres_2023}: given an image and a referring expression, a model must return one, many, or zero target instances, with outputs evaluated as bounding boxes or instance masks.

We introduce \textbf{AgroVG}, a multi-source agricultural visual grounding benchmark with 10{,}071 annotation-grounded image--query pairs drawn from 10 source datasets and spanning six target families: crop/weed, fruit, wheat head, pest, plant disease, and tree canopy. AgroVG defines two complementary tasks (Fig.~\ref{fig:teaser}): T1 (bounding-box grounding) for coarse localization across all six target families, and T2 (instance-mask grounding) for pixel-precise localization on the subset of families with reliable instance-level pixel annotations. Both tasks are evaluated under single-target, multi-target, and target-absent regimes, yielding six diagnostic cells. A central principle is \emph{annotation-grounded query construction}: every positive query is traceable to a verified target set and specific object identities, while every target-absent query is verified to have an empty target set under the benchmark annotations. To make AgroVG a rigorous testbed, we normalize heterogeneous source annotations into a unified instance-level schema, apply expert quality control, and pair the data with task-specific protocols for box-set matching and query-level mask coverage, detailed in Sec.~\ref{sec:evaluation}.

Zero-shot evaluation of 26 model configurations spanning closed-source MLLMs, open-source VLMs, and specialized grounding systems shows that AgroVG is far from saturated: even the best T1 multi-target configuration reaches only 0.35 Set-$F_1$, and the best T2 positive-query mask success rate remains below 0.17 at IoU@0.75. Models also show poor abstention calibration, either hallucinating boxes or masks on target-absent queries or over-abstaining on positive queries. These results position AgroVG as a challenging and diagnostic benchmark for agricultural vision-language grounding.

In summary, our contributions are:

\begin{itemize}[leftmargin=*, itemsep=0.20em, topsep=0.20em, parsep=0pt, partopsep=0pt]
    \item We formulate agricultural visual grounding as a generalized set-prediction problem and provide task-specific evaluation protocols for bounding-box grounding (T1) and instance-mask grounding (T2) across single-target, multi-target, and target-absent regimes.

    \item We construct AgroVG, a multi-source, multi-family benchmark with 10{,}071 annotation-grounded image--query pairs and verified object identities, covering six agricultural target families under quality-controlled annotations.

    \item We conduct a zero-shot evaluation of 26 representative configurations spanning closed-source MLLMs, open-source VLMs, and specialized grounding systems, revealing systematic failures in set completeness, existence-aware abstention, and fine-grained mask grounding.
\end{itemize}

\begin{table*}[!t]
\centering
\setlength{\belowcaptionskip}{4pt}
\caption{
\textbf{Benchmark comparison.}
AgroVG uniquely combines multi-source agricultural coverage with box and mask grounding across single-, multi-, and target-absent settings.
}
\label{tab:benchmark_comparison}

\begingroup
\scriptsize
\setlength{\tabcolsep}{3.0pt}
\renewcommand{\arraystretch}{0.96}
\setlength{\aboverulesep}{0.25ex}
\setlength{\belowrulesep}{0.25ex}
\setlength{\cmidrulesep}{0.15ex}

\resizebox{\textwidth}{!}{%
\begin{tabular}{l c c l c l | c c c | c c c}
\toprule
\multirow{2}{*}{\textbf{Benchmark}}
& \multirow{2}{*}{\textbf{Year}}
& \multirow{2}{*}{\textbf{Dom.}}
& \multirow{2}{*}{\textbf{Size}}
& \multirow{2}{*}{\textbf{\#Src.}}
& \multirow{2}{*}{\textbf{Target}}
& \multicolumn{3}{c|}{\textbf{BBox Grounding}}
& \multicolumn{3}{c}{\textbf{Mask Grounding}} \\
\cmidrule(lr){7-9} \cmidrule(lr){10-12}
& & & & &
& \textbf{Single} & \textbf{Multi} & \textbf{Absent}
& \textbf{Single} & \textbf{Multi} & \textbf{Absent} \\
\midrule

RefCOCO~\cite{yu_modeling_2016}
& 2016 & Nat. & 19{,}994 img. & 1 & general
& \cmark & \xmark & \xmark
& \cmark & \xmark & \xmark \\

gRefCOCO~\cite{liu_gres_2023}
& 2023 & Nat. & 19{,}994 img. & 1 & general
& \xmark & \xmark & \xmark
& \cmark & \cmark & \cmark \\

\midrule

DIOR-RSVG~\cite{zhan_rsvg_2023}
& 2023 & RS & 17{,}402 img. & 1 & aerial
& \cmark & \xmark & \xmark
& \xmark & \xmark & \xmark \\

RRSIS-D~\cite{liu_rotated_2024}
& 2024 & RS & 17{,}402 trip. & 1 & aerial
& \xmark & \xmark & \xmark
& \cmark & \xmark & \xmark \\

VRSBench~\cite{ding_vrsbench_2024}
& 2024 & RS & 29{,}614 img. & 2 & aerial
& \cmark & \xmark & \xmark
& \xmark & \xmark & \xmark \\

\midrule

AgriBench~\cite{zhou_agribench_2024}
& 2024 & Agri. & 1{,}784 img. & 1 & --
& \xmark & \xmark & \xmark
& \xmark & \xmark & \xmark \\

AgroBench~\cite{shinoda_agrobench_2025}
& 2025 & Agri. & 4{,}218 img. & multi & --
& \xmark & \xmark & \xmark
& \xmark & \xmark & \xmark \\

AgroMind~\cite{li_can_2025}
& 2025 & Agri-RS & 20{,}850 img. & 10 & --
& \xmark & \xmark & \xmark
& \xmark & \xmark & \xmark \\

AgMMU~\cite{gauba_agmmu_2025}
& 2025 & Agri. & -- & -- & --
& \xmark & \xmark & \xmark
& \xmark & \xmark & \xmark \\

\cmidrule(lr){1-12}

gRef-CW~\cite{haghighat_multilabel_2026}
& 2026 & Agri. & 8{,}034 img. & 1 & crop/weed
& \cmark & \cmark & \cmark
& \xmark & \xmark & \xmark \\

\rowcolor{gray!18}
\textbf{AgroVG (Ours)}
& \textbf{2026} & \textbf{Agri.}
& \textbf{10{,}071 pairs}
& \textbf{10}
& \textbf{6 families}
& \cmark & \cmark & \cmark
& \cmark & \cmark & \cmark \\

\bottomrule
\end{tabular}%
}

\vspace{0.25em}
\begin{minipage}{0.98\textwidth}
\footnotesize
\emph{Notes.} Size follows each benchmark's reported unit: img. = images, trip. = triplets, and pairs = image--query pairs.
Target denotes the domain of language-conditioned localization; ``--'' indicates no instance-level grounding protocol.
AgroVG covers six agricultural target families: crop/weed, fruit, wheat head, pest, plant disease, and tree canopy.
\end{minipage}

\endgroup
\end{table*}

\section{Related Work}
\label{sec:related_work}

\subsection{Visual Grounding Benchmarks}
Visual grounding connects natural-language expressions to localized visual evidence and is commonly studied along two complementary axes: Referring Expression Comprehension (REC) with bounding boxes and Referring Expression Segmentation (RES) with masks~\cite{hu_segmentation_2016,yu_mattnet_2018,deng_transvg_2021}. Natural-image benchmarks such as RefCOCO, RefCOCO+~\cite{yu_modeling_2016}, and RefCOCOg~\cite{mao_generation_2016} established canonical single-target grounding protocols on everyday objects, where each expression is typically assumed to refer to exactly one target. More recent generalized grounding benchmarks, such as gRefCOCO and the GREC/GRES settings~\cite{he_grec_2023,liu_gres_2023}, relax this assumption by introducing multi-target and target-absent expressions, reformulating grounding as a set-prediction problem. Together, these efforts establish two evaluation axes, output granularity (box vs.\ mask) and target-set cardinality (single, multi, absent), that we adopt and instantiate for agricultural imagery in AgroVG.

Domain-specific grounding benchmarks have also emerged in remote sensing. \emph{Box-level grounding} benchmarks include RSVG~\cite{sun_visual_2022}, DIOR-RSVG\cite{zhan_rsvg_2023}, and subsequent datasets that extend referring expression comprehension to higher-resolution, optical, and drone-view imagery~\cite{lan_language_2024,li_languageguided_2024,liu_aerialvg_2025,sun_refdrone_2025}. \emph{Mask-level referring segmentation} benchmarks extend language-conditioned localization to pixel-precise outputs~\cite{yuan_rrsis_2024,dong_crossmodal_2025,yang_largescale_2025,ye_rislad_2026}. More recent work further scales this line toward large-scale, generalized, and reasoning-oriented remote-sensing grounding~\cite{quenum_lisat_2025,li_segearthr1_2025,xin_segearthr2_2025,jiang_grasp_2025,ni_unigeoseg_2026}, as well as spatial and cross-view reasoning~\cite{chu_natural_2025,zhou_geoground_2025}. VRSBench~\cite{ding_vrsbench_2024} additionally combines object-level references with VQA and captioning. Despite this rapid progress, the remote-sensing regime, typically top-down imagery dominated by vehicles, ships, buildings, aircraft, and land-cover regions, differs fundamentally from close-range agricultural imagery in scale distributions, target densities, occlusion patterns, and viewpoint characteristics. This leaves agricultural visual grounding systematically under-evaluated.

\subsection{Agricultural Vision-Language Benchmarks}
Recent agricultural vision-language benchmarks have advanced the evaluation of multimodal models in agricultural settings~\cite{yang_agrigpt_2025,yang_agrigptomni_2025,yang_agrigptvl_2025,yan_agrieval_2026}. AgriBench~\cite{zhou_agribench_2024}, AgroBench~\cite{shinoda_agrobench_2025}, AgroMind~\cite{li_can_2025}, and AgMMU~\cite{gauba_agmmu_2025} evaluate agricultural vision-language models on tasks such as recognition, visual question answering, captioning, and domain-specific reasoning. While these benchmarks are important for measuring high-level agricultural understanding, they generally do not provide a unified instance-level protocol for language-conditioned localization. A model may correctly answer whether a disease is present or describe a fruit scene, but this does not test whether it can localize the exact diseased regions, retrieve all valid target instances, or abstain when the referred target is absent.

Agricultural detection and segmentation datasets provide rich object-level supervision for crop--weed segmentation, fruit detection, plant phenotyping, pest recognition, disease segmentation, and tree canopy analysis. However, standard detection and segmentation are not equivalent to visual grounding: they predict instances from predefined categories, whereas grounding requires selecting the target set specified by a natural-language expression~\cite{yu_modeling_2016}. The same image may require one instance, multiple instances, or no output depending on the query. The closest related agricultural grounding benchmark is gRef-CW~\cite{haghighat_multilabel_2026}, which provides box-level generalized referring expression comprehension for CropAndWeed imagery but no mask-grounding evaluation. As discussed in Sec.\ref{sec:intro} and summarized in Table~\ref{tab:benchmark_comparison}, AgroVG complements this line of work along three dimensions: multi-source aggregation, multi-family coverage, and joint bounding-box and instance-mask evaluation.

\section{AgroVG Benchmark}
\label{agrovg}

\begin{figure*}[!t]
  \centering
  \setlength{\abovecaptionskip}{3pt}
  \includegraphics[width=0.97\textwidth]{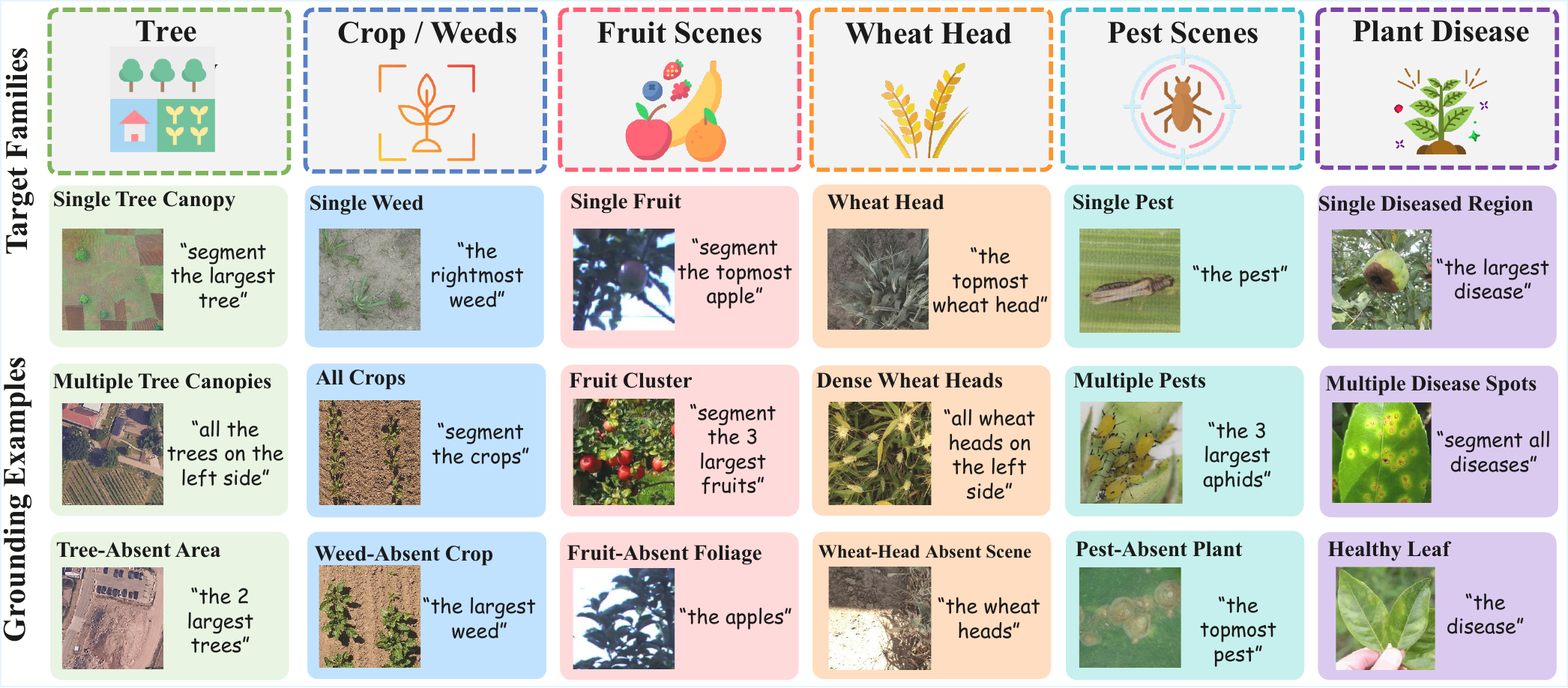}
  \caption{
  \textbf{Target families and grounding regimes in AgroVG.}
  AgroVG covers six target families (columns): tree canopy, 
  crop/weed, fruit, wheat head, pest, and plant disease. For each family, 
  we show queries under three regimes (rows): 
  \emph{single-target}, \emph{multi-target}, and \emph{target-absent}, 
  where the output is one localized instance, a set of instances, 
  or an empty prediction.
  }
  \label{fig:target_examples}
\end{figure*}

\subsection{Task Formulation}
\label{sec:task_formulation}
AgroVG formulates agricultural visual grounding as a \emph{generalized set-prediction problem}. Given an image $I$ and a referring expression $q$, the goal is to localize all visual targets satisfying $q$. Let $\mathcal{O}(I)$ be the annotated instances in $I$. Each query induces a target set $\mathcal{G}(I,q) \subseteq \mathcal{O}(I)$ with $K = |\mathcal{G}(I,q)|$, where $K$ is determined by both the image content and the query rather than fixed a priori. AgroVG therefore supports single-target, multi-target, and target-absent grounding under two complementary output protocols: bounding-box grounding (T1) and instance-mask grounding (T2).

\noindent\textbf{T1: Bounding-box grounding.}
T1 represents each target instance by an axis-aligned box. The ground-truth output is $\mathcal{B}^{*}(I,q) = \{b(o) \mid o \in \mathcal{G}(I,q)\}$, where $b(o) = (x^{\min}, y^{\min}, x^{\max}, y^{\max})$, and a model predicts a box set $\hat{\mathcal{B}}(I,q)$. T1 evaluates box-level localization and aligns with the output format of grounding detectors and many grounding-capable MLLMs.

\noindent\textbf{T2: Instance-mask grounding.}
T2 represents each target instance by a binary mask. The ground-truth output is $\mathcal{M}^{*}(I,q) = \{m(o) \mid o \in \mathcal{G}(I,q)\}$, where $m(o) \in \{0,1\}^{H \times W}$, and a model predicts a mask set $\hat{\mathcal{M}}(I,q)$. T2 evaluates pixel-precise grounding for agricultural targets with irregular shapes, partial occlusion, or dense overlap.

\noindent\textbf{Query regimes.}
Following generalized grounding settings~\cite{liu_gres_2023,he_grec_2023}, 
AgroVG evaluates three regimes shared by T1 and T2: \emph{single-target} 
queries with $K=1$, \emph{multi-target} queries with $K>1$, and 
\emph{target-absent} queries with $K=0$, where no verified target instance 
within the benchmark target space satisfies the expression. These regimes 
probe object localization, set completeness, and existence-aware 
abstention, yielding six $\langle\text{task}, \text{regime}\rangle$ 
evaluation cells. Fig.\ref{fig:target_examples} illustrates these 
regimes across the six target families with representative referring 
queries.

\subsection{Benchmark Construction}
\label{sec:benchmark_construction}

\begin{figure*}[!t]
  \centering
  \setlength{\abovecaptionskip}{3pt}
  \includegraphics[width=0.97\textwidth]{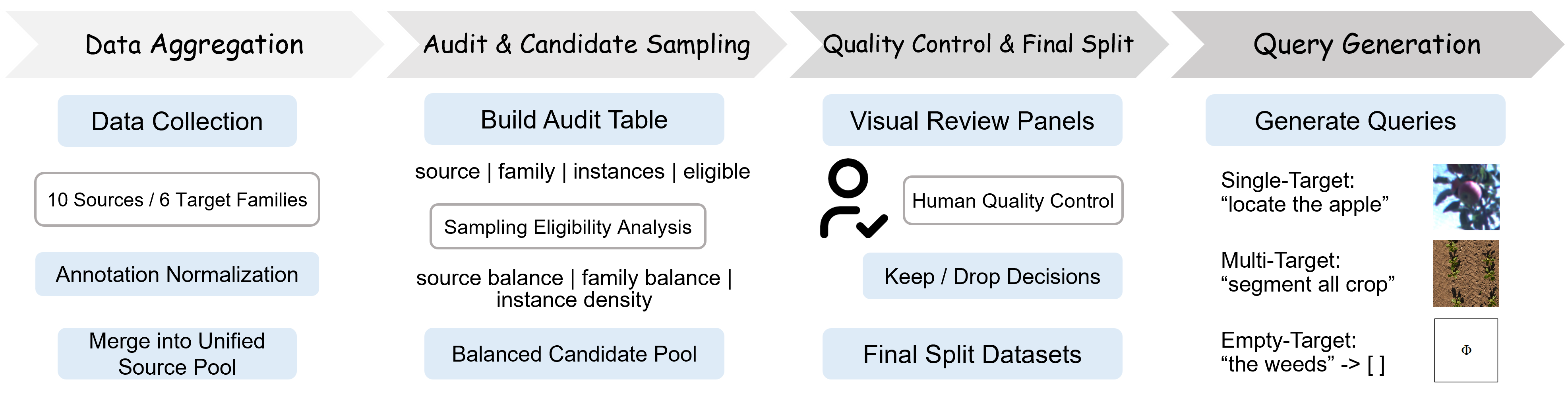}
\caption{
\textbf{AgroVG construction pipeline.}
AgroVG converts heterogeneous agricultural datasets into verified image--query pairs through annotation normalization, audit-driven sampling, expert visual review, split assignment, and split-aware query generation. Positive and target-absent queries are generated after image verification, ensuring each query is tied to stable ground-truth annotations.
}
  \label{fig:pipeline}
\end{figure*}

As shown in Fig.~\ref{fig:pipeline}, AgroVG follows a split-aware construction order: candidate images are first normalized, audited, reviewed, and assigned to final splits before referring query generation. This ensures each query is derived from a verified image and stable ground-truth annotations.

\noindent\textbf{Data sources and annotation normalization.}
AgroVG is built from ten source datasets covering six target families: \emph{crop/weed} (PhenoBench~\cite{weyler_phenobench_2024}, CropAndWeed~\cite{steininger_cropandweed_2023}), \emph{fruit} (MinneApple~\cite{hani_minneapple_2020}, ACFR~\cite{bargoti_deep_2017}, MegaFruits~\cite{wang_learn_2026}), \emph{wheat head} (GWHD~\cite{david_global_2021}), \emph{pest} (IP102~\cite{wu_ip102_2019}), \emph{plant disease} (PlantSeg~\cite{wei_largescale_2026}), and \emph{tree canopy} (OAM-TCD~\cite{veitch-michaelis_oamtcd_2024}, Dumortier et al.~\cite{dumortier_annotated_2025}). T1 uses nine sources (all except MegaFruits) covering all six target families, while T2 is instantiated from five mask-capable sources (PhenoBench, CropAndWeed, PlantSeg, MegaFruits, ACFR) covering three families: crop/weed, fruit, and plant disease; a per-source overview is provided in Table~\ref{tab:source_datasets}. We normalize all sources into a unified instance-level schema with image metadata, target family, instance ID, bounding box, mask reference when available, and annotation provenance. For T2, masks are used only when supported by pixel-level evidence. Native instance masks are used directly when available; otherwise, instance masks are derived from pixel-level annotations combined with instance cues, such as circle seeds, stem points, or polygons. AgroVG never synthesizes masks from bounding-box geometry alone. Detailed derivation rules and filtering thresholds are provided in Appendix~\ref{app:schema_and_masks}.

\noindent\textbf{Audit-driven sampling and expert quality control.}
The ten sources vary substantially in scale and target density. To prevent imbalance, we construct an image-level \emph{audit table} recording source identity, target family, queryable instance counts, density buckets, and eligibility for single-target, multi-target, and target-absent queries. Candidate sampling is controlled along three axes: \emph{source-aware sampling}, \emph{family-level balancing}, and \emph{instance-density stratification}, with multi-ready images preferred for multi-target query construction. Audit fields, source quotas, and density-bucket statistics are detailed in Appendix~\ref{app:sampling_qc}. Each candidate image is rendered as a review panel by overlaying normalized boxes or masks on the original RGB image, and ten domain experts screen candidates to remove corrupted images, invalid annotations, mask misalignment, ambiguous targets, and visually unreliable cases. The retained images are partitioned into \texttt{dev} and \texttt{test} splits at the image level, stratified by source dataset and target family, with no image appearing in both splits.

\noindent\textbf{Split-aware query generation.}
Queries are generated only from the approved final splits, ensuring that each query is linked to a verified image and a stable ground-truth target set. Each query record stores the image identity, query text, task type, query regime, and the associated target set: T1 records store target object IDs and target boxes, while T2 records store target instance IDs used to recover the query mask; for target-absent queries, the target set is empty. Queries are annotation-grounded and template-based, covering category-level and fine-grained references, geometric rank (largest, leftmost, etc.), spatial side selection, relative relations between target families, set-level quantification, and target-absent negatives. For T1, queries are target descriptions whose expected outputs are bounding-box sets; for T2, queries are segmentation-oriented expressions whose expected outputs are instance-mask sets. Generated queries are automatically verified for target-count consistency and ambiguity, and a stratified sample from each $\langle\text{source}, \text{regime}\rangle$ cell undergoes additional manual inspection. The full template library and verification rules are provided in Appendix~\ref{app:query_generation}.

\begin{figure*}[!t]
    \centering
    \setlength{\abovecaptionskip}{3pt}
    \includegraphics[width=0.99\textwidth]{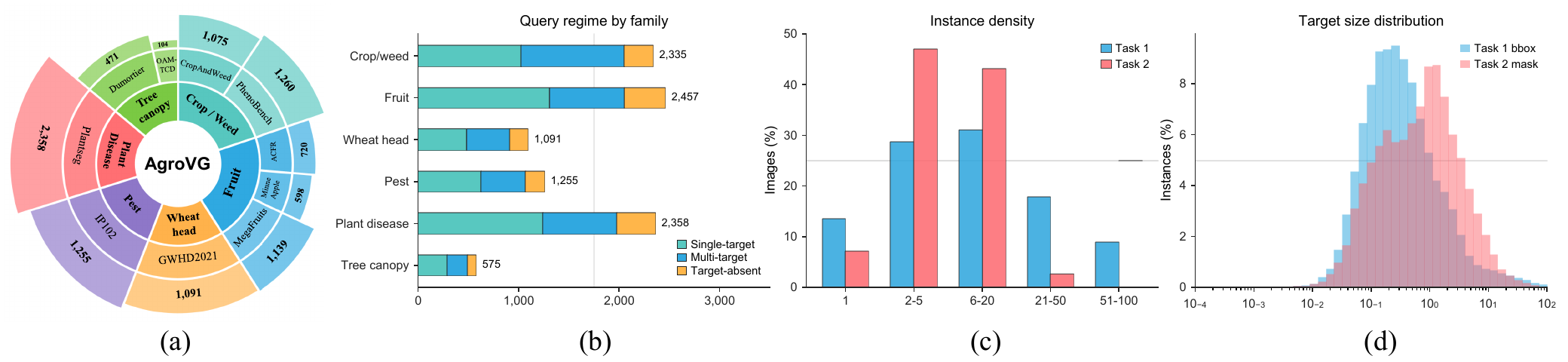}
\caption{
\textbf{Dataset statistics of AgroVG.}
Panels summarize source/family composition, query-regime counts, instance-density buckets, and target-size distributions for T1 and T2.
}
    \label{fig:statistics}
\end{figure*}

\noindent\textbf{Dataset statistics.}
After the pipeline, AgroVG comprises $10{,}071$ annotation-grounded image--query pairs over expert-reviewed agricultural images. T1 covers all six target families, while T2 is instantiated on the three families supported by reliable instance-level pixel annotations (crop/weed, fruit, and plant disease). As shown in Fig.~\ref{fig:statistics}, source contributions are bounded so that no single source or family dominates (a); each family preserves a meaningful target-absent share alongside positive queries (b); and both T1 and T2 cover a wide instance-density and target-size range, with most examples in moderately dense scenes of 2--20 instances per image (c,d), reflecting the multi-target nature of agricultural grounding. Compared with existing agricultural grounding resources, AgroVG uniquely combines multi-source and multi-family coverage with both bounding-box and instance-mask grounding under single-target, multi-target, and target-absent regimes.

\section{Evaluation Protocol}
\label{sec:evaluation}
We evaluate AgroVG under three principles consistent with its generalized grounding formulation: 
(i) \emph{box-set matching} for T1, 
(ii) \emph{query-level mask coverage} for T2, and 
(iii) \emph{multi-granularity reporting} across query regimes, sources, target families, and program types. 
We define metrics separately for the two output protocols because T1 predicts instance-level bounding-box sets, whereas T2 predicts query-level binary masks induced by target instance IDs.

\subsection{T1: Bounding-Box Set Matching}
\label{sec:eval_t1}
For each T1 query, the model prediction is parsed into a set of valid boxes $\hat{\mathcal{B}}(I,q)$ and compared with the ground-truth box set $\mathcal{B}^{*}(I,q)$. For each IoU threshold $\tau \in \{0.50,0.75\}$, we construct a bipartite graph between ground-truth and predicted boxes, adding an edge whenever the pairwise box IoU is at least $\tau$. We then compute a maximum-cardinality matching on this thresholded graph. The objective is to maximize the number of threshold-satisfying matches rather than the sum of matched IoUs.

Matched pairs are counted as $\mathrm{TP}_{\tau}$, unmatched predictions as $\mathrm{FP}_{\tau}$, and unmatched ground-truth boxes as $\mathrm{FN}_{\tau}$. We compute query-level precision, recall, and Set-$F_1$, and report Set-$F_1$ at $\tau \in \{0.50,0.75\}$ as the primary T1 metric. We report macro aggregation as the primary T1 score, which averages query-level scores; micro aggregation is included in the released evaluator summaries.

For single-target queries, we additionally report a strict accuracy: a query is correct only when the model predicts exactly one valid box and that box matches the unique ground-truth box. For target-absent queries, the correct prediction is the empty set. Empty predictions are counted as correct for Empty Accuracy, while any predicted box on a target-absent query is counted as a false positive.

\subsection{T2: Query-Level Mask Coverage}
\label{sec:eval_t2}
For T2, the ground-truth mask is generated from the image-level instance map and the query's target instance IDs. Pixels belonging to any target instance are foreground, yielding a query-level binary mask. The model output is also evaluated as a binary mask after thresholding. Although T2 is constructed from instance-level annotations, evaluation measures query-level coverage of the referred target region rather than per-instance mask matching, avoiding the need to compare heterogeneous predictors by instance-level mask outputs.

For each positive query, we compute pixel-level intersection, union, ground-truth area, and predicted area, from which IoU and Dice are derived. We report query-averaged mIoU and mDice, as well as cumulative cIoU and cDice computed by summing intersections, unions, and foreground areas across positive queries before taking the ratio. mIoU treats each query equally, while cIoU reflects overall pixel-level coverage and is more influenced by large masks. We also report query-level success rates IoU@0.50 and IoU@0.75, using thresholded success rather than AP-style metrics because the evaluated models produce confidence scores with different calibrations or no confidence scores at all. For target-absent queries, the ground-truth mask is empty; an empty prediction is counted as correct for Empty Accuracy, while any predicted foreground is counted in the empty false-positive rate.

\subsection{Reporting Granularity}
\label{sec:eval_reporting}
Combining two output protocols and three query regimes yields six $\langle\text{task},\text{regime}\rangle$ evaluation cells. Main results report the primary metrics for each cell: strict accuracy for T1 single-target queries, Set-$F_1$ at IoU thresholds for T1 multi-target queries, Empty Accuracy for target-absent queries, and mIoU/cIoU/IoU@$ \tau $ for T2 positive queries. Each cell is broken down by source dataset, target family, query program type, and, for T2, mask provenance. Full precision, recall, Set-$F_1$, Dice, false-positive rates, and diagnostic breakdowns at the standard thresholds are provided in Appendix~\ref{app:metrics}.

\begin{table*}[!t]
\centering
\setlength{\belowcaptionskip}{4pt}
\footnotesize
\setlength{\tabcolsep}{3pt}
\renewcommand{\arraystretch}{0.96}

\caption{
Main results on \textbf{AgroVG-T1}, grouped by model category.
All/Fam denote overall/family-macro Set-$F_1$; S-Acc, M-F1, and E-Acc are single-, multi-, and empty-query diagnostics.
\textbf{Bold} and \underline{underlined} values indicate the best and second-best results.
}
\label{tab:t1_main_results}

\begin{tabularx}{\textwidth}{@{}>{\raggedright\arraybackslash}p{0.22\textwidth}*{6}{Y}@{}}
\toprule
\multirow{2}{*}{\textbf{Model}}
& \multicolumn{3}{c}{\textbf{Set-$F_1$}}
& \multicolumn{3}{c}{\textbf{Diagnostics}} \\
\cmidrule(lr){2-4}
\cmidrule(lr){5-7}
& \textbf{All@.5}
& \textbf{Fam@.5}
& \textbf{All@.75}
& \textbf{S-Acc}
& \textbf{M-F1}
& \textbf{E-Acc} \\
\midrule

\rowcolor{gray!15}
\multicolumn{7}{@{}l}{\textbf{\textit{Closed-source MLLMs}}} \\
GPT-4o~\cite{openai_hello_2024}
& 0.1505 & 0.1512 & 0.1234 & 0.0451 & 0.0146 & 0.7990 \\
GPT-5.4~\cite{openai_gpt54_2026}
& 0.2736 & 0.2612 & 0.1421 & 0.2287 & 0.1653 & 0.6683 \\
Gemini 2.5 Pro~\cite{geminiteam_gemini_2023}
& 0.1849 & 0.1744 & 0.0793 & 0.1401 & 0.1583 & 0.3769 \\
Claude Sonnet 4.6~\cite{anthropic_introducing_2026}
& 0.1995 & 0.1900 & 0.0997 & 0.1691 & 0.1132 & 0.4975 \\

\midrule

\rowcolor{gray!15}
\multicolumn{7}{@{}l}{\textbf{\textit{Open-source VLMs}}} \\
{\footnotesize DeepSeek-VL2-tiny}~\cite{wu_deepseekvl2_2024}
& 0.2227 & 0.2153 & 0.1817 & 0.1031 & 0.0616 & \textbf{0.9749} \\
{\footnotesize DeepSeek-VL2-small}~\cite{wu_deepseekvl2_2024}
& 0.3191 & 0.3057 & 0.2179 & 0.2995 & 0.2344 & 0.5528 \\
InternVL3.5-8B~\cite{wang_internvl35_2025}
& \underline{0.3963} & \underline{0.3816} & 0.2387 & \underline{0.3430} & \textbf{0.3540} & 0.5829 \\
InternVL3.5-38B~\cite{wang_internvl35_2025}
& \textbf{0.3978} & \textbf{0.3920} & \textbf{0.2569} & \textbf{0.3527} & \underline{0.2980} & 0.7588 \\
Qwen3-VL-8B~\cite{bai_qwen3vl_2025}
& 0.3271 & 0.3174 & 0.2186 & 0.2415 & 0.1737 & 0.8894 \\
Qwen3-VL-32B~\cite{bai_qwen3vl_2025}
& 0.3748 & 0.3673 & \underline{0.2393} & 0.3188 & 0.2626 & 0.8090 \\

\midrule

\rowcolor{gray!15}
\multicolumn{7}{@{}l}{\textbf{\textit{Specialized grounding systems}}} \\
OWLv2-Large~\cite{minderer_scaling_2023}
& 0.2951 & 0.2819 & 0.2361 & 0.0805 & 0.1576 & \underline{0.9246} \\
Florence-2-Large~\cite{xiao_florence2_2024}
& 0.2019 & 0.1871 & 0.1326 & 0.2576 & 0.2034 & 0.0000 \\
{\footnotesize Grounding DINO 1.5}~\cite{liu_grounding_2025}
& 0.2471 & 0.2286 & 0.1659 & 0.1417 & 0.2910 & 0.0151 \\

\bottomrule
\end{tabularx}

\end{table*}

\section{Experiments}
\label{sec:experiments}
We benchmark representative models on AgroVG under zero-shot evaluation: no model is trained or fine-tuned on AgroVG. Our goal is to characterize how existing multimodal and grounding systems generalize to agricultural visual grounding, rather than to optimize a model specifically for the benchmark. The evaluation spans three model classes that differ in training paradigm, accessibility, and grounding interface: \emph{closed-source MLLMs}, \emph{open-source VLMs}, and \emph{specialized grounding systems}. All models are evaluated on the AgroVG \texttt{test} split under the protocols defined in Sec.~\ref{sec:evaluation}; the development split is used only for prompt selection, output-format design, and threshold tuning.

\subsection{Evaluated Models and Setup}
\label{sec:exp_setup}
\noindent\textbf{Evaluated models.}
We evaluate 26 task-specific configurations on AgroVG: 4 closed-source MLLMs, 6 open-source VLMs, and 16 specialized grounding configurations. The closed-source models are GPT-4o~\cite{openai_hello_2024}, GPT-5.4~\cite{openai_gpt54_2026}, Gemini 2.5 Pro~\cite{geminiteam_gemini_2023}, and Claude Sonnet 4.6~\cite{anthropic_introducing_2026}. The open-source VLMs cover three families at two scales: DeepSeek-VL2 (tiny, small)~\cite{wu_deepseekvl2_2024}, InternVL3.5 (8B, 38B)~\cite{wang_internvl35_2025}, and Qwen3-VL (8B, 32B)~\cite{bai_qwen3vl_2025}, all used for T1 box outputs. Specialized configurations include T1 grounding detectors, OWLv2-Large~\cite{minderer_scaling_2023}, Florence-2-Large~\cite{xiao_florence2_2024}, and Grounding DINO 1.5~\cite{liu_grounding_2025}; T2 segmentation MLLMs, LISA (7B, 13B)~\cite{lai_lisa_2024}, GLaMM~\cite{rasheed_glamm_2024}, PixelLM (7B, 13B)~\cite{ren_pixellm_2024}, GSVA (force-seg)~\cite{xia_gsva_2024}, and PSALM~\cite{zhang_psalm_2025}; and T2 promptable/two-stage text-to-mask methods, X-Decoder~\cite{zou_generalized_2023}, SEEM~\cite{zou_segment_2023}, Florence-2-Large, SAM~3~\cite{carion_sam_2025}, Grounded-SAM~\cite{ren_grounded_2024}, and OWLv2-Large+SAM.  Two-stage methods are marked with +SAM in Table~\ref{tab:t2_main_results}. Open-source and specialized models run on NVIDIA A100 GPUs, while closed-source models are queried via official APIs.

\begin{table*}[!t]
\centering
\setlength{\belowcaptionskip}{4pt}
\footnotesize
\setlength{\tabcolsep}{3pt}
\renewcommand{\arraystretch}{0.96}

\caption{
Main results on \textbf{AgroVG-T2}, grouped by mask-grounding interface.
All/Fam denote overall/family-macro mask scores; IoU@ thresholds are positive-query success rates, and E-Acc is empty-query accuracy.
\textbf{Bold} and \underline{underlined} values indicate the best and second-best results.
}
\label{tab:t2_main_results}

\begin{tabularx}{\textwidth}{@{}>{\raggedright\arraybackslash}p{0.27\textwidth}*{6}{Y}@{}}
\toprule
\multirow{2}{*}{\textbf{Model}}
& \multicolumn{3}{c}{\textbf{Mask Quality}}
& \multicolumn{3}{c}{\textbf{Success / Abstention}} \\
\cmidrule(lr){2-4}
\cmidrule(lr){5-7}
& \textbf{All-mIoU}
& \textbf{Fam-mIoU}
& \textbf{cIoU}
& \textbf{IoU@.50}
& \textbf{IoU@.75}
& \textbf{E-Acc} \\
\midrule

\rowcolor{gray!15}
\multicolumn{7}{@{}l}{\textbf{\textit{Segmentation MLLMs / reasoning segmentation models}}} \\
LISA-7B~\cite{lai_lisa_2024}
& \underline{0.3088} & \underline{0.3020} & 0.3869 & \textbf{0.3085} & 0.1542 & 0.0283 \\
LISA-13B~\cite{lai_lisa_2024}
& \textbf{0.3177} & \textbf{0.3131} & 0.3879 & \textbf{0.3085} & \underline{0.1575} & 0.0755 \\
GLaMM~\cite{rasheed_glamm_2024}
& 0.2628 & 0.2540 & 0.2905 & 0.2206 & 0.1161 & 0.0377 \\
PixelLM-7B~\cite{ren_pixellm_2024}
& 0.3080 & 0.2870 & 0.3829 & 0.2886 & 0.1111 & 0.0000 \\
PixelLM-13B~\cite{ren_pixellm_2024}
& 0.2903 & 0.2706 & \underline{0.3890} & 0.2687 & 0.1078 & 0.0189 \\
GSVA (force-seg)~\cite{xia_gsva_2024}
& 0.2159 & 0.2037 & 0.2252 & 0.1808 & 0.0896 & 0.0000 \\
PSALM~\cite{zhang_psalm_2025}
& 0.2980 & 0.2759 & \textbf{0.4151} & \underline{0.2935} & \textbf{0.1675} & 0.0000 \\

\midrule

\rowcolor{gray!15}
\multicolumn{7}{@{}l}{\textbf{\textit{Promptable / universal text-to-mask models}}} \\
X-Decoder~\cite{zou_generalized_2023}
& 0.1470 & 0.1337 & 0.2313 & 0.1376 & 0.0796 & 0.3868 \\
SEEM~\cite{zou_segment_2023}
& 0.1868 & 0.1739 & 0.2290 & 0.1857 & 0.1012 & 0.2925 \\
Florence-2-Large~\cite{xiao_florence2_2024}
& 0.1875 & 0.1603 & 0.1866 & 0.1360 & 0.1078 & 0.1981 \\
SAM 3~\cite{carion_sam_2025}
& 0.1161 & 0.0941 & 0.3112 & 0.0995 & 0.0680 & \textbf{0.9340} \\

\midrule

\rowcolor{gray!15}
\multicolumn{7}{@{}l}{\textbf{\textit{Two-stage grounding-to-mask references}}} \\
Grounded-SAM~\cite{ren_grounded_2024}
& 0.2450 & 0.2430 & 0.2843 & 0.2322 & 0.1360 & 0.0094 \\
{\footnotesize OWLv2-Large+SAM}~\cite{carion_sam_2025}
& 0.1042 & 0.1149 & 0.0825 & 0.0912 & 0.0498 & \underline{0.6792} \\

\bottomrule
\end{tabularx}

\end{table*}

\noindent\textbf{Experimental setup.}
For T1, each model receives an image and a referring expression and is instructed to return all matching target instances as a list of axis-aligned bounding boxes; for target-absent queries, the required output is an empty list. For T2, models with native mask outputs are queried directly using segmentation-oriented expressions, while two-stage methods pass predicted boxes to SAM as box prompts to obtain final masks. We use the same task-level prompt template whenever the model interface permits, with only minimal model-specific formatting adjustments. All outputs are processed by a unified pipeline that normalizes box coordinates to image-pixel space when needed, clips boxes to image bounds, discards malformed predictions, and scores predictions with the official AgroVG evaluation scripts. We use greedy or deterministic decoding whenever supported; otherwise, the provider default is recorded. Beyond these pre-specified parsing and validity rules, we do not apply test-set post-hoc filtering, re-ranking, ensembling, or manual correction. Full model lists, version identifiers, prompt templates, parsing rules, decoding settings, confidence thresholds, and hardware details are provided in Appendix~\ref{app:models_prompts}.

\noindent\textbf{T1: Bounding-box grounding.}
Table~\ref{tab:t1_main_results} reports the main results on AgroVG-T1.
Open-source VLMs achieve the strongest box-grounding performance among the evaluated model groups, with InternVL3.5-38B leading on the primary metrics and stricter localization, while InternVL3.5-8B performs best on multi-target Set-$F_1$.
This contrast suggests that stronger overall localization does not necessarily imply better set completion.
Scaling improves both Qwen3-VL and DeepSeek-VL2, although the gains remain uneven across diagnostic axes.

\begin{figure*}[!t]
  \centering
  \setlength{\abovecaptionskip}{3pt}
  \includegraphics[width=0.98\textwidth]{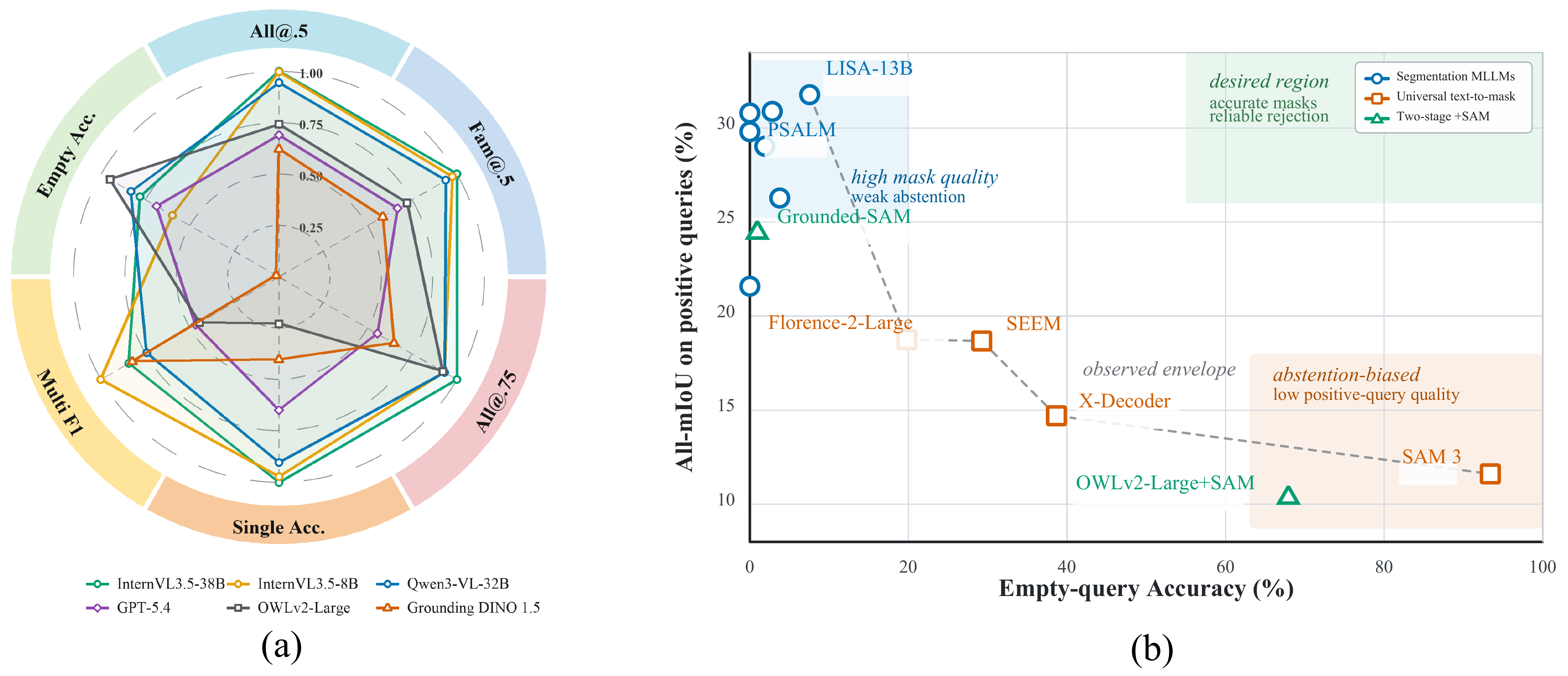}
  \caption{
  \textbf{Zero-shot diagnostics on AgroVG.}
  (a) T1 models across localization, set-completeness, and abstention; (b) T2 trade-off between positive-query mask quality and empty-query accuracy.
  }
  \label{fig:results}
\end{figure*}

Closed-source MLLMs and specialized grounding systems exhibit complementary failure modes rather than uniform dominance.
GPT-5.4 is the strongest closed-source model but still trails InternVL3.5-38B, whereas GPT-4o shows stronger target-absent rejection but weaker positive-target localization.
Among specialized systems, OWLv2-Large abstains well on target-absent queries, while Grounding DINO 1.5 and Florence-2-Large favor positive detections but abstain poorly.
As shown in Fig.\ref{fig:results}(a), no single model dominates all T1 axes, confirming that AgroVG-T1 jointly tests localization accuracy, target-set completeness, and calibrated abstention.

\noindent\textbf{T2: Instance-mask grounding.}
Table~\ref{tab:t2_main_results} summarizes the main results on AgroVG-T2.
Segmentation MLLMs and reasoning segmentation models produce the strongest positive-query masks, with LISA-13B leading in query-averaged and family-macro mIoU, and PSALM showing stronger cumulative coverage and stricter mask success. Nevertheless, success rates remain low, indicating that pixel-level agricultural grounding remains challenging for dense, irregular, and partially occluded targets.

The main T2 distinction is the trade-off between positive-query mask quality and target-absent rejection. SAM~3 achieves the strongest target-absent rejection but weak positive-query mask quality, whereas LISA-style and other mask-oriented models segment positive targets better but often hallucinate foreground when the target is absent. Two-stage systems show the same tension: Grounded-SAM improves positive-query mask coverage but rarely abstains, while OWLv2-Large+SAM abstains more reliably but produces weaker masks. As shown in Fig.~\ref{fig:results}(b), no model reaches the desired region where high positive-query mask quality coexists with reliable target-absent rejection.

\noindent\textbf{Key findings.}
Across both tasks, AgroVG exposes three persistent gaps in current models. First, strong positive localization does not guarantee set completeness: the best T1 multi-target Set-$F_1$ is only 0.3540, indicating that models often miss valid target instances in dense agricultural scenes. Second, abstention is poorly calibrated: models with high Empty accuracy often behave conservatively, while models with stronger positive localization frequently hallucinate boxes or masks when the queried target is absent. Third, mask grounding remains the hardest protocol: the best T2 All-mIoU stays below 0.32 and the best IoU@.75 success rate remains below 0.17. These findings suggest that reliable agricultural grounding requires joint progress on target existence, target cardinality, and fine-grained geometric coverage.

\section{Conclusion}
\label{sec:conclusion}
In this work, we introduce \textbf{AgroVG}, a multi-source benchmark for agricultural visual grounding. AgroVG contains 10{,}071  expert-verified pairs from ten source datasets, covers six target families, and supports bounding-box grounding (T1) and instance-mask grounding (T2) under single-target, multi-target, and target-absent settings. With annotation-grounded queries and task-specific evaluation protocols, AgroVG provides a standardized testbed for diagnosing agricultural grounding models. Our zero-shot evaluation of 26 model configurations reveals persistent gaps in multi-target completeness, existence-aware abstention, and pixel-level mask grounding for dense or irregular targets. These results highlight the need for grounding models that are both localization-accurate and existence-aware in complex agricultural scenes. AgroVG is currently limited by template-based referring expressions and by the subset of source datasets that supply reliable instance-level pixel annotations, which restricts T2 to three of the six target families. Future work will expand query diversity, incorporate additional mask-capable sources, and study controlled adaptation of grounding models to agricultural imagery.

\bibliographystyle{unsrtnat}
\bibliography{agrovg}


\appendix

\renewcommand{\thesection}{\Alph{section}}
\renewcommand{\thesubsection}{\Alph{section}.\arabic{subsection}}
\renewcommand{\thefigure}{A\arabic{figure}}
\renewcommand{\thetable}{A\arabic{table}}
\setcounter{figure}{0}
\setcounter{table}{0}

\clearpage
\thispagestyle{empty}

\begin{center}
{\Large\bfseries AgroVG: A Large-Scale Multi-Source Benchmark for Agricultural Visual Grounding\par}
\vspace{0.3em}
{\large\bfseries (Supplementary Material)\par}
\end{center}

\vspace{1.8em}

\noindent{\large\bfseries Table of Contents in Appendix}

\vspace{1.0em}

\newcommand{\appsecentry}[2]{%
    \noindent\textbf{#1}\dotfill \pageref{#2}\par\vspace{0.35em}
}

\newcommand{\appsubentry}[2]{%
    \noindent\hspace*{1.8em}#1\dotfill \pageref{#2}\par\vspace{0.20em}
}

{\normalsize
\appsecentry{A \quad Broader Impacts}{app:broader_impact}

\appsecentry{B \quad Dataset Construction Details}{app:dataset_construction}
\appsubentry{B.1 \quad Source Datasets, Task Coverage, and Licenses}{app:source_datasets}
\appsubentry{B.2 \quad Annotation Schema and Mask Derivation}{app:schema_and_masks}
\appsubentry{B.3 \quad Sampling, Quality Control, and Splits}{app:sampling_qc}
\appsubentry{B.4 \quad Query Generation and Dataset Statistics}{app:query_generation}

\appsecentry{C \quad Evaluation Setup and Reproducibility}{app:evaluation_setup}
\appsubentry{C.1 \quad Metric Definitions and Edge Cases}{app:metrics}
\appsubentry{C.2 \quad Evaluated Models, Prompts, and Parsing}{app:models_prompts}
\appsubentry{C.3 \quad Compute and Reproducibility Artifacts}{app:compute}

\appsecentry{D \quad Additional Results and Diagnostics}{app:additional_results}
\appsubentry{D.1 \quad T1 Diagnostic Results}{app:extended_t1}
\appsubentry{D.2 \quad T2 Diagnostic Results}{app:extended_t2}
\appsubentry{D.3 \quad Failure Taxonomy and Qualitative Examples}{app:qualitative}

\appsecentry{E \quad Datasheet, Hosting, Licensing, and Maintenance}{app:hosting}
\appsecentry{F \quad Case Study}{app:case_studies}
}

\clearpage

\section{Broader Impacts}
\label{app:broader_impact}

AgroVG is intended to support research on reliable language-conditioned localization in plant-centric agricultural imagery. By aggregating ten source datasets across crop/weed, fruit, wheat head, pest, plant disease, and tree canopy targets, and by evaluating bounding-box and instance-mask grounding under single-target, multi-target, and target-absent regimes, AgroVG provides a standardized testbed for capabilities that are important for precision agriculture, instruction-following agricultural robots, selective harvesting, targeted weed control, and early disease or pest monitoring. Its diagnostic metrics highlight not only localization accuracy, but also multi-target completeness, existence-aware abstention, and pixel-level mask quality.

The benchmark also surfaces failure modes that are important to identify before deployment. Our zero-shot evaluation shows that current models remain weak in multi-target set completion, abstention calibration, and fine-grained mask grounding. In real agricultural systems, such failures could lead to incorrect interventions, such as treating healthy plants as diseased, missing parts of a weed cluster, or acting on a target that is absent from the observed scene. AgroVG should therefore be viewed as an evaluation benchmark rather than evidence of deployment readiness. Models evaluated on AgroVG require additional validation under the target crop variety, sensor configuration, geographic region, and field condition before use in operational decision-making.

Potential misuse should also be considered. Although AgroVG contains no human subjects, no facial imagery, and no personally identifiable information, improvements in generalized grounding could transfer to monitoring applications outside agriculture, including human or property surveillance. The AgroVG release is therefore intended for agricultural research use, and should follow the licenses, access restrictions, and intended-use conditions of the contributing source datasets. As with other agricultural automation technologies, downstream applications may also affect labor demand in scouting, weeding, harvesting, and crop monitoring; such impacts should be considered alongside technical performance, especially in regions where agricultural labor markets are sensitive to mechanization.

To mitigate these risks, we encourage users to report disaggregated results by source dataset, target family, and query regime; to validate models with additional regional data before deployment; and to avoid treating AgroVG scores as a certification of readiness for autonomous agricultural intervention. The benchmark is best used as a scientific tool for developing more reliable, instruction-aware, and existence-aware agricultural grounding systems.

\section{Dataset Construction Details}
\label{app:dataset_construction}

\begin{figure}[t]
  \centering
  \includegraphics[
    width=0.98\textwidth,
    height=0.78\textheight,
    keepaspectratio
  ]{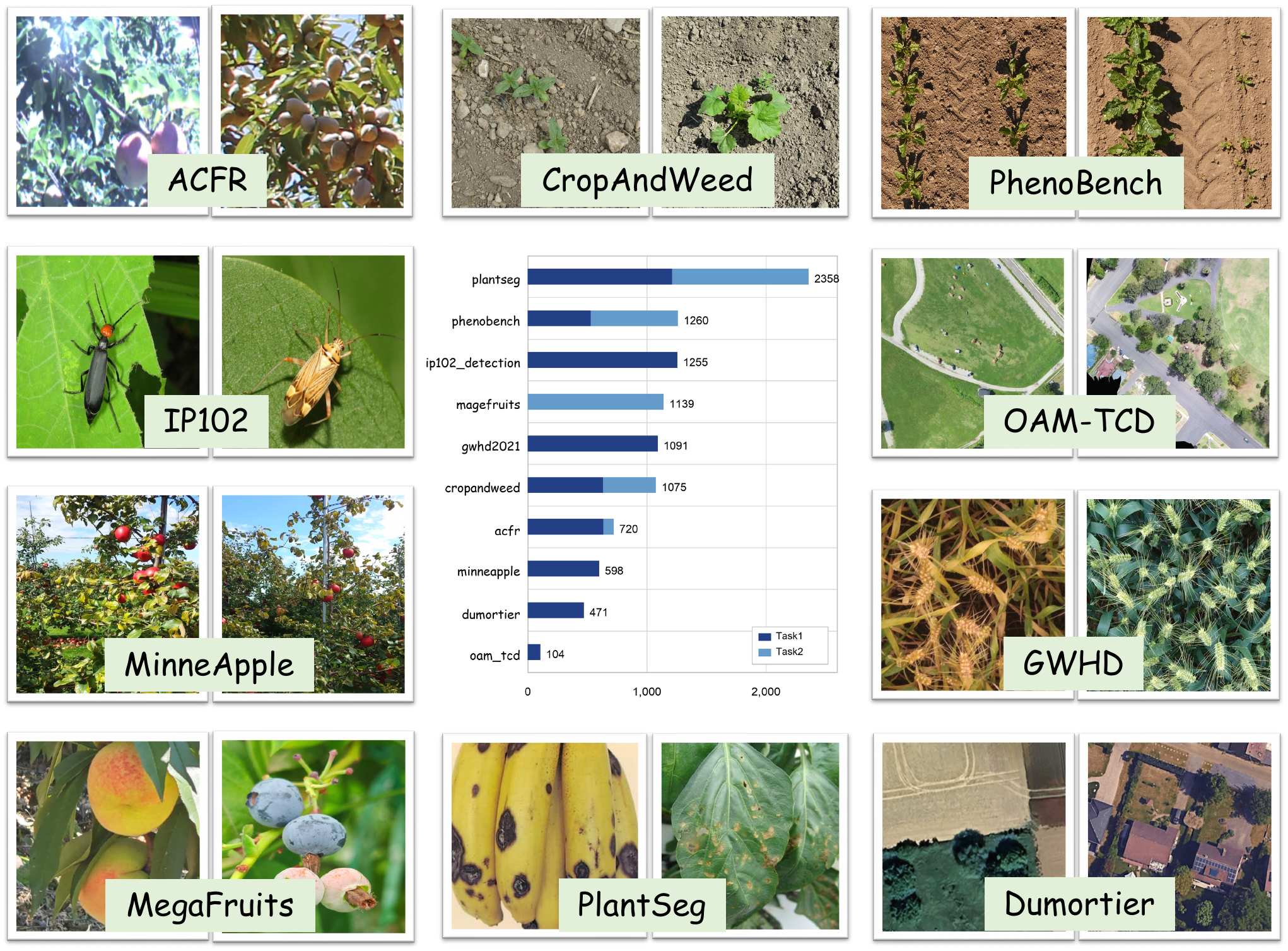}
  \caption{
  \textbf{Source datasets in AgroVG.}
  Representative examples from the ten agricultural source datasets are shown around the center plot.
  The center bar chart reports the number of image--query pairs contributed by each source, split by T1 bounding-box grounding and T2 instance-mask grounding.
  }
  \label{fig:appendix_source_overview}
\end{figure}

\subsection{Source Datasets, Task Coverage, and Licenses}
\label{app:source_datasets}

AgroVG aggregates ten public agricultural datasets covering six target families: crop/weed, fruit, wheat head, pest, plant disease, and tree canopy. We did not collect new imagery. Instead, we convert mature agricultural detection and segmentation resources into a unified grounding benchmark with annotation-grounded referring queries. This design improves reproducibility and allows AgroVG to cover a broader set of agricultural subdomains than a newly collected single-source dataset.

Sources are selected when they satisfy three conditions: they depict plant-centric agricultural targets, provide instance-level boxes or pixel-level evidence from which instance-level supervision can be obtained, and are publicly accessible under terms compatible with academic benchmarking. T1 uses sources with reliable instance-level localizations and covers all six target families. T2 imposes the stricter requirement of consistent instance-level pixel evidence under the AgroVG normalization protocol; MinneApple and OAM-TCD are used only for T1 in the current release because their segmentation annotations were not included in the audited T2 pool, which is a coverage limitation rather than a fundamental restriction. Table~\ref{tab:source_datasets} summarizes source coverage, annotation evidence, task usage, and access information.

AgroVG does not override source-dataset licenses. Each record retains source provenance and attribution. When source terms restrict raw-image redistribution, AgroVG releases metadata, preprocessing scripts, and query annotations, and users obtain raw images from the original source. Full per-source license statements are provided in the released dataset card.

\subsection{Annotation Schema and Mask Derivation}
\label{app:schema_and_masks}
The ten source datasets use different category vocabularies, coordinate conventions, annotation formats, and metadata structures. AgroVG normalizes them into a unified instance-level representation with three record layers (image, instance, query) summarized in Table~\ref{tab:agrovg_schema_compact}; the full JSONL schema is released with the dataset metadata. T1 target sets are represented by object IDs and boxes; T2 target sets additionally store instance-map IDs; target-absent queries store an empty target set.

Bounding boxes use absolute-pixel \texttt{xyxy} coordinates with 
the origin at the upper-left image corner; T2 masks use uint16 instance maps aligned with the native image grid. After normalization, all records are subjected to automated validity checks, including valid dimensions, non-degenerate boxes, in-bound coordinates, non-empty masks, and box--mask consistency where applicable. Records failing these checks are removed or excluded from query generation.

\begin{table}[t]
\centering
\setlength{\belowcaptionskip}{4pt}
\caption{
\textbf{Source datasets in AgroVG.}
For each source we report the target family, native annotation evidence, AgroVG task usage, mask provenance for T2, native resolution, and access location. \emph{Derived} masks are produced by the AgroVG normalization pipeline from source-side pixel evidence and instance cues.
}
\label{tab:source_datasets}

\scriptsize
\setlength{\tabcolsep}{3pt}
\renewcommand{\arraystretch}{1.10}

\resizebox{\textwidth}{!}{%
\begin{tabular}{@{}l c l l c c l l l@{}}
\toprule
\textbf{Source} & \textbf{Year} & \textbf{Family} & \textbf{Native annotation} &
\textbf{T1} & \textbf{T2} & \textbf{Mask prov.} & \textbf{Resolution} & \textbf{Access} \\
\midrule
PhenoBench~\cite{weyler_phenobench_2024}            & 2024 & crop/weed     & instance masks                & \cmark & \cmark & native   & 1024$\times$1024 & \scriptsize\url{phenobench.org} \\
CropAndWeed~\cite{steininger_cropandweed_2023}      & 2023 & crop/weed     & boxes + semantic + stem pts.  & \cmark & \cmark & derived  & high-res         & \scriptsize\url{github.com/cropandweed} \\
MinneApple~\cite{hani_minneapple_2020}              & 2020 & fruit         & polygonal masks               & \cmark & ---    & ---      & high-res         & \scriptsize\url{rsn.umn.edu/MinneApple} \\
ACFR~\cite{bargoti_deep_2017}                       & 2016 & fruit         & foreground + circle seeds     & \cmark & \cmark & derived  & 308$\times$202   & \scriptsize\url{data.acfr.usyd.edu.au} \\
MegaFruits~\cite{wang_learn_2026}                   & 2026 & fruit         & instance masks                & ---    & \cmark & native   & variable         & \scriptsize\url{kaggle.com/mmwang0} \\
GWHD 2021~\cite{david_global_2021}                  & 2021 & wheat head    & bounding boxes                & \cmark & ---    & ---      & 1024$\times$1024 & \scriptsize\url{global-wheat.com} \\
IP102~\cite{wu_ip102_2019}                          & 2019 & pest          & bounding boxes                & \cmark & ---    & ---      & variable         & \scriptsize\url{github.com/xpwu95/IP102} \\
PlantSeg~\cite{wei_largescale_2026}                & 2026 & plant disease & polygons + binary mask        & \cmark & \cmark & derived  & variable         & \scriptsize\url{zenodo.org/13762907} \\
OAM-TCD~\cite{veitch-michaelis_oamtcd_2024}      & 2024 & tree canopy   & instance + semantic masks     & \cmark & ---    & ---      & 2048$\times$2048 & \scriptsize\url{zenodo.org/11617167} \\
Dumortier~et al.~\cite{dumortier_annotated_2025}    & 2025 & tree canopy   & instance annotations          & \cmark & ---    & ---      & 640$\times$640   & \scriptsize\url{zenodo.org/15155081} \\
\bottomrule
\end{tabular}%
}
\end{table}

\begin{table}[t]
\centering
\caption{
\textbf{Conceptual AgroVG annotation schema.}
The released metadata contains the full field-level schema; this table summarizes the three record layers used for grounding evaluation.
}
\label{tab:agrovg_schema_compact}

\small
\setlength{\tabcolsep}{6pt}
\renewcommand{\arraystretch}{1.15}
\begin{tabularx}{\textwidth}{@{}l X X@{}}
\toprule
\textbf{Layer} & \textbf{Essential information} & \textbf{Purpose} \\
\midrule
Image    
& Source, split, image size, group ID, RGB path, and T2 instance-map path when available.
& Source traceability and leakage-aware split control. \\

Instance 
& Object ID, target family, source class, bounding box, mask ID when available, and annotation provenance.
& Defines localizable agricultural targets with auditable origin. \\

Query    
& Query text, task, regime, program type, target IDs, target boxes, and T2 target mask IDs.
& Defines the annotation-grounded answer set for grounding evaluation. \\
\bottomrule
\end{tabularx}
\end{table}

\paragraph{Mask provenance.}
AgroVG-T2 uses native instance masks when available and derives masks only when the source provides pixel-level evidence. PhenoBench and MegaFruits provide native instance masks and are normalized directly. ACFR, CropAndWeed, and PlantSeg require source-specific derivation from pixel-level evidence and instance cues. AgroVG never synthesizes masks by filling bounding boxes, since box-filled masks would overestimate irregular agricultural targets and confound pixel-level evaluation. Each derived instance stores its mask provenance in the released metadata. All native and derived masks pass source-specific minimum-area and minimum-side filters (Table~\ref{tab:mask_thresholds}) before entering the T2 pool, with thresholds reflecting source-specific target scales.

\paragraph{ACFR-derived masks.}
ACFR provides binary fruit foreground masks together with circle annotations for individual fruit instances. We treat the binary foreground as the mask domain and use circle centers as instance seeds. Marker-based watershed on the distance transform partitions the foreground into instance regions. When a seed falls outside the foreground, it is projected to the nearest foreground pixel; when foreground or seed annotations are incomplete, we use conservative fallback rules based on annotation disks or connected components. Images without usable pixel-level fruit evidence are excluded from T2 rather than approximated from boxes.

\paragraph{CropAndWeed-derived masks.}
CropAndWeed provides semantic crop/weed masks, bounding boxes, and stem-point annotations. We process each semantic class independently: the class-specific foreground defines the mask domain, while stem points serve as instance seeds, with box-center fallback when a stem point is outside the foreground. Marker-based watershed partitions the semantic region into candidate instances. To reduce leakage between neighboring plants, each candidate is trimmed by its corresponding box region and reduced to the connected component containing the seed. This produces instance masks grounded in semantic pixels and instance cues rather than box geometry alone.

\paragraph{PlantSeg-derived masks.}
PlantSeg provides polygon annotations for disease regions, together with binary disease-foreground masks. We rasterize valid polygons into instance maps and preserve source split consistency during normalization. When polygon annotations are unavailable but a non-empty binary disease mask exists, we fall back to connected components on the binary foreground. If neither usable polygons nor binary foreground are available, the image is dropped from T2. Thus PlantSeg masks are derived from polygonal or pixel-level disease evidence, never from bounding boxes.

\begin{table}[t]
\centering
\caption{
\textbf{T2 mask provenance, derivation rules, and filtering thresholds.}
$A_{\min}$ is the minimum mask area in pixels; $S_{\min}$ is the minimum side length of the tight mask box. Derived masks are produced from pixel-level evidence and instance cues, never by filling bounding boxes.
}
\label{tab:mask_thresholds}

\small
\setlength{\tabcolsep}{4pt}
\renewcommand{\arraystretch}{1.20}
\begin{tabularx}{\textwidth}{@{}l l X c c@{}}
\toprule
\textbf{Source} & \textbf{Provenance} & \textbf{Normalization or derivation rule} & $\boldsymbol{A_{\min}}$ & $\boldsymbol{S_{\min}}$ \\
\midrule
PhenoBench~\cite{weyler_phenobench_2024}
& Native
& Native plant instance masks are normalized and filtered.
& 64 & 36 \\

MegaFruits~\cite{wang_learn_2026}
& Native
& Native fruit instance masks are normalized and filtered.
& 16 & 4 \\

ACFR~\cite{bargoti_deep_2017}
& Derived
& Binary fruit foreground is partitioned using circle-center seeds.
& 8 & 2 \\

CropAndWeed~\cite{steininger_cropandweed_2023}
& Derived
& Semantic crop/weed foreground is partitioned using stem-point or box-center cues, followed by cleanup.
& 24 & 36 \\

PlantSeg~\cite{wei_largescale_2026}
& Derived
& Disease-region polygons are rasterized into instance masks; connected-component fallback is used only when polygons are unavailable but binary foreground exists.
& 4 & 2 \\
\bottomrule
\end{tabularx}

\vspace{0.35em}
\begin{minipage}{\textwidth}
\footnotesize
\raggedright
Instances failing either threshold are excluded from the T2 pool. Rejection counts are reported in the released per-source statistics.
\end{minipage}
\end{table}

\subsection{Sampling, Quality Control, and Splits}
\label{app:sampling_qc}

The source datasets differ substantially in scale, target density, annotation provenance, and original task formulation. Uniform sampling from the merged source pool would over-represent the largest contributors and under-represent low-density, multi-target, and target-absent regimes. AgroVG therefore applies audit-driven sampling and visual review before query generation.

\paragraph{Audit table and candidate sampling.}
Source-side normalization outputs are merged into a unified pool with globally unique image and instance identifiers, while preserving source-side group identifiers (e.g., repeated captures of the same plant) for leakage-aware split construction. For each image, AgroVG computes an audit row recording source identity, target family, queryable instance count, instance-density bucket, mask provenance when applicable, and eligibility for single-target, multi-target, and target-absent query construction. A balanced T2 review pool of $N_{\text{review}}=5{,}000$ images is drawn from this audit table under three controls: source-aware quotas (preventing source dominance and approximately balancing the three T2 families), multi-ready preference (ensuring sufficient multi-target query coverage), and density stratification (preserving sparse, moderate, and dense scenes). The resulting quotas and retained counts are reported in Table~\ref{tab:audit_quotas}.

\paragraph{Expert visual review.}
Each sampled image is rendered as an RGB-overlay panel before query generation. T1 panels overlay normalized boxes, while T2 panels overlay instance masks and boundaries. A panel of ten graduate students with backgrounds in remote sensing, land science, and computer science reject images with corrupted imagery, visibly incomplete annotations, severe box or mask misalignment, merged or over-split instances, or RGB--annotation mismatch. The review focuses on geometric and visual annotation integrity rather than taxonomic relabeling; source-provided category labels are retained. Each image is reviewed once, and we therefore do not report inter-reviewer agreement; retained and dropped image IDs are released to support future audits.

\paragraph{Split construction.}
Review-approved images form the final image pools. For T2, $3{,}545$ of the $5{,}000$ reviewed candidates are retained. The retained images are partitioned into \texttt{dev} and \texttt{test} splits with an approximately $80\%/20\%$ ratio, applied per source. When source metadata defines groups, all images sharing a \texttt{group\_id} are assigned to the same split to avoid leakage. Query generation is performed only after split assignment and is described in Appendix~\ref{app:query_generation}.

\begin{table}[t]
\centering
\caption{
\textbf{T2 sampling and quality-control summary.}
Review quota is the number of sampled candidate images before visual review. Retained images form the final T2 pool. Pass rates reflect source-specific differences in target density, annotation provenance, and mask reliability.
}
\label{tab:audit_quotas}

\small
\setlength{\tabcolsep}{5pt}
\renewcommand{\arraystretch}{1.15}

\begin{tabularx}{\textwidth}{@{}l l r r r X@{}}
\toprule
\textbf{Source} & \textbf{Family} & \textbf{Review quota} & \textbf{Retained} & \textbf{Pass rate} & \textbf{Preferred density buckets} \\
\midrule
PhenoBench           & crop/weed     & 900   & 732   & 81.3\% & $\{1, \text{2--5}, \text{6--20}, \text{21--50}\}$ \\
CropAndWeed          & crop/weed     & 800   & 443   & 55.4\% & $\{1, \text{2--5}, \text{6--20}, \text{21--50}\}$ \\
MegaFruits-blueberry & fruit         & 700   & 554   & 79.1\% & $\{\text{2--5}, \text{6--20}, \text{21--50}\}$ \\
MegaFruits-peach     & fruit         & 700   & 585   & 83.6\% & $\{\text{2--5}, \text{6--20}, \text{21--50}\}$ \\
ACFR-apples          & fruit         & 300   & 84    & 28.0\% & $\{1, \text{2--5}, \text{6--20}, \text{21--50}\}$ \\
PlantSeg             & plant disease & 1{,}600 & 1{,}147 & 71.7\% & $\{1, \text{2--5}, \text{6--20}, \text{21--50}\}$ \\
\midrule
\textbf{Total}       & ---           & \textbf{5{,}000} & \textbf{3{,}545} & \textbf{70.9\%} & --- \\
\bottomrule
\end{tabularx}
\end{table}

\subsection{Query Generation and Dataset Statistics}
\label{app:query_generation}
Queries are generated only after image normalization, visual review, and split assignment. This split-aware order ensures that each released query is tied to stable ground-truth annotations and that no query crosses dev/test image boundaries. AgroVG comprises $10{,}071$ released image--query pairs ($6{,}526$ for T1 and $3{,}545$ for T2), with query-type budgets of approximately $50\%$ 
single-target, $35\%$ multi-target, and $15\%$ target-absent. T2 follows a one-query-per-image policy on its $3{,}545$ final images, yielding a $1{:}1$ correspondence; T1 releases $6{,}526$ selected eligible image--query pairs from a larger expert-verified image pool of $6{,}664$ images.

\paragraph{Design principles.}
The template library is governed by three principles. \emph{(i)
Annotation-grounded:} every emitted query carries an explicit target
set traceable to verified instance annotations; positive queries store
target object IDs and boxes (T1) or instance IDs (T2), and target-absent
queries are verified to have an empty target set under the global
family space. \emph{(ii) Deterministic and program-typed:} each query
is produced by exactly one named template (its \texttt{program\_type},
stored in the released records), enabling per-program-type evaluation
breakdowns; rank candidates that fail separation margins are dropped
rather than emitted as ambiguous expressions. \emph{(iii) Stratified:} candidates are distributed across target families within each query-type bucket using largest-remainder allocation, ensuring that no single family dominates within a bucket. T1 additionally separates multi-target queries into core and challenge sub-buckets to balance simpler family-all queries against more discriminative top-$K$ and side-selection queries.

\paragraph{Selection and verification.}
For each final-pool image, the generator enumerates all template candidates and runs automatic checks for target-count consistency, eligibility-flag agreement, and absent-family validity. Selection follows the budget reported above, with multi-target candidates ranked by quality and density preferences. A stratified subsample across $\langle\text{source}, \text{query\_type}\rangle$ cells is drawn for manual inspection; queries failing manual review are excluded from the released set. Table~\ref{tab:query_templates} lists the complete template library used in the released benchmark.

\begin{table}[t]
\centering
\caption{
\textbf{AgroVG query template library.}
Each program type defines a triggering condition and a target-set resolution rule. The \emph{Group} column gives a reader-friendly aggregation used in diagnostic analyses, while \emph{Program type} is the exact identifier stored in the released query records.
}
\label{tab:query_templates}

\scriptsize
\setlength{\tabcolsep}{3pt}
\renewcommand{\arraystretch}{1.12}

\begin{tabularx}{\textwidth}{@{}l l l c X X@{}}
\toprule
\textbf{Regime} & \textbf{Group} & \textbf{Program type} & \textbf{Tasks} & \textbf{Trigger condition} & \textbf{Target resolution} \\
\midrule
Single & Unique & \texttt{single\_fine\_unique} & T1, T2
& Fine class has exactly one instance.
& The unique instance. \\

Single & Unique & \texttt{single\_family\_unique} & T1, T2
& Family has exactly one queryable instance.
& The unique instance. \\

Single & Spatial rank & \texttt{single\_rank\_leftmost} & T1, T2
& Family has $\geq 2$ queryable instances; T1 requires separation.
& Instance with smallest bbox-center $x$. \\

Single & Spatial rank & \texttt{single\_rank\_rightmost} & T1, T2
& Family has $\geq 2$ queryable instances; T1 requires separation.
& Instance with largest bbox-center $x$. \\

Single & Spatial rank & \texttt{single\_rank\_topmost} & T1, T2
& Family has $\geq 2$ queryable instances; T1 requires separation.
& Instance with smallest bbox-center $y$. \\

Single & Spatial rank & \texttt{single\_rank\_bottommost} & T1, T2
& Family has $\geq 2$ queryable instances; T1 requires separation.
& Instance with largest bbox-center $y$. \\

Single & Size rank & \texttt{single\_rank\_largest} & T1, T2
& Family has $\geq 2$ queryable instances; T1 requires area separation.
& Largest bbox area for T1; largest mask area for T2. \\

Single & Size rank & \texttt{single\_rank\_smallest} & T2
& Family has $\geq 2$ queryable instances.
& Instance with smallest mask area. \\

Single & Relation & \texttt{single\_relation\_family} & T1
& Crop and weed coexist with a unique anchor relation.
& The unique target left/right/above/below the anchor. \\

\midrule

Multi & All & \texttt{multi\_family\_all} & T1, T2
& Family has $\geq 2$ queryable instances and passes the count guardrail.
& All queryable instances of that family. \\

Multi & Top-$K$ & \texttt{multi\_family\_topk\_leftmost} & T1, T2
& Family has a separated top-$K$ set.
& Top-$K$ by smallest bbox-center $x$. \\

Multi & Top-$K$ & \texttt{multi\_family\_topk\_largest} & T1, T2
& Family has a separated top-$K$ set.
& Top-$K$ by largest area. \\

Multi & Side subset & \texttt{multi\_family\_side\_left} & T1
& Valid left-side subset with at least two targets.
& All queryable instances in the left-side subset. \\

Multi & Side subset & \texttt{multi\_family\_side\_right} & T1
& Valid right-side subset with at least two targets.
& All queryable instances in the right-side subset. \\

Multi & Fine-class all & \texttt{multi\_fine\_all} & T1
& Fine class has $\geq 2$ queryable instances.
& All queryable instances of that fine class. \\

Multi & Fine-class top-$K$ & \texttt{multi\_fine\_topk\_leftmost} & T1
& Fine class has a separated top-$K$ set.
& Top-$K$ by smallest bbox-center $x$ within the fine class. \\

Multi & Fine-class top-$K$ & \texttt{multi\_fine\_topk\_largest} & T1
& Fine class has a separated top-$K$ set.
& Top-$K$ by bbox area within the fine class. \\

Multi & Fine-class side & \texttt{multi\_fine\_side\_right} & T1
& Realized fine-class right-side subset.
& All queryable fine-class instances in the right-side subset. \\

\midrule

Empty & Absent family & \texttt{empty\_family\_negative} & T1
& Target family is absent from queryable instances.
& Empty set. \\

Empty & Absent family & \texttt{empty\_family\_absent} & T2
& Target family is absent from queryable instances.
& Empty set. \\

\bottomrule
\end{tabularx}
\end{table}

\begin{table}[t]
\centering
\caption{
\textbf{Per-source dev/test composition.}
For each source, we report the number of image-query pairs in the dev and test splits and the total. The realized dev fraction stays close to the target $80\%$ across all sources, reflecting the source-stratified split policy in Appendix~\ref{app:sampling_qc}.
}
\label{tab:per_source_split}

\setlength{\tabcolsep}{6pt}
\renewcommand{\arraystretch}{1.15}

\begin{tabular}{@{}l l r r r c@{}}
\toprule
\textbf{Task} & \textbf{Source} & \textbf{Dev} & \textbf{Test} & \textbf{Total} & \textbf{Dev frac.} \\
\midrule
T1 & ACFR              & 509   & 127   & 636   & 80.0\% \\
T1 & CropAndWeed       & 505   & 127   & 632   & 79.9\% \\
T1 & Dumortier         & 377   & 94    & 471   & 80.0\% \\
T1 & GWHD 2021         & 889   & 202   & 1{,}091 & 81.5\% \\
T1 & IP102             & 1{,}004 & 251   & 1{,}255 & 80.0\% \\
T1 & MinneApple        & 480   & 118   & 598   & 80.3\% \\
T1 & OAM-TCD           & 83    & 21    & 104   & 79.8\% \\
T1 & PhenoBench        & 422   & 106   & 528   & 79.9\% \\
T1 & PlantSeg          & 969   & 242   & 1{,}211 & 80.0\% \\
\midrule
\multicolumn{2}{@{}l}{\textbf{T1 subtotal}} & \textbf{5{,}238} & \textbf{1{,}288} & \textbf{6{,}526} & \textbf{80.3\%} \\
\midrule
T2 & ACFR              & 67    & 17    & 84    & 79.8\% \\
T2 & CropAndWeed       & 354   & 89    & 443   & 79.9\% \\
T2 & MegaFruits-blueberry & 443 & 111   & 554   & 80.0\% \\
T2 & MegaFruits-peach     & 468 & 117   & 585   & 80.0\% \\
T2 & PhenoBench        & 586   & 146   & 732   & 80.1\% \\
T2 & PlantSeg          & 918   & 229   & 1{,}147 & 80.0\% \\
\midrule
\multicolumn{2}{@{}l}{\textbf{T2 subtotal}} & \textbf{2{,}836} & \textbf{709} & \textbf{3{,}545} & \textbf{80.0\%} \\
\midrule
\multicolumn{2}{@{}l}{\textbf{AgroVG total}} & \textbf{8{,}074} & \textbf{1{,}997} & \textbf{10{,}071} & \textbf{80.2\%} \\
\bottomrule
\end{tabular}
\end{table}

\paragraph{Per-source composition.}
The $80\%/20\%$ dev/test split ratio is preserved at the per-source 
level (Table~\ref{tab:per_source_split}), ensuring that source-level evaluation breakdowns remain meaningful on the test set. The joint source--target-family distribution (Table~\ref{tab:source_family_pivot}) exhibits a diagonal-plus-off-diagonal structure: diagonal cells carry positive queries grounded in the source's primary family, while off-diagonal cells carry target-absent queries derived from verified family absence rather than synthetic mismatches. This structure enables existence-aware abstention evaluation without cross-source query mixing.

\begin{table}[t]
\centering
\caption{
\textbf{Per-source $\times$ per-target-family image-query pair counts.}
Diagonal cells (positive queries grounded in the source's primary family) carry the bulk of each source's contribution; off-diagonal cells correspond to target-absent queries asking the model to abstain on families absent from the image. Empty cells are zero.
}
\label{tab:source_family_pivot}

\small
\setlength{\tabcolsep}{4pt}
\renewcommand{\arraystretch}{1.15}

\begin{tabular}{@{}l l r r r r r r r r@{}}
\toprule
\textbf{Task} & \textbf{Source} & \textbf{Crop} & \textbf{Disease} & \textbf{Fruit} & \textbf{Pest} & \textbf{Tree} & \textbf{Weed} & \textbf{Wheat head} & \textbf{Total} \\
\midrule
T1 & ACFR             & 9   & 5    & \textbf{613}  & 9    &       &      &     & 636   \\
T1 & CropAndWeed      & \textbf{219} & 35   & 29   & 21   &       & \textbf{328} &     & 632   \\
T1 & Dumortier        &     & 23   & 25   & 19   & \textbf{404} &      &     & 471   \\
T1 & GWHD 2021        &     & 66   & 45   & 71   &       &      & \textbf{909} & 1{,}091 \\
T1 & IP102            & 64  & 52   & 72   & \textbf{1{,}067} &       &      &     & 1{,}255 \\
T1 & MinneApple       & 57  & 48   & \textbf{435}  & 58   &       &      &     & 598   \\
T1 & OAM-TCD          &     & 6    & 7    & 6    & \textbf{85}   &      &     & 104   \\
T1 & PhenoBench       & \textbf{275} & 32   & 27   & 30   &       & \textbf{164} &     & 528   \\
T1 & PlantSeg         & 73  & \textbf{1{,}030} & 61   & 47   &       &      &     & 1{,}211 \\
\midrule
\multicolumn{2}{@{}l}{\textbf{T1 total}} & 697 & 1{,}297 & 1{,}314 & 1{,}328 & 489 & 492 & 909 & \textbf{6{,}526} \\
\midrule
T2 & ACFR             & 5   & 5    & \textbf{71}   &      &       & 3    &     & 84    \\
T2 & CropAndWeed      & \textbf{186} & 14   & 33   &      &       & \textbf{210} &     & 443   \\
T2 & MegaFruits-blueberry & 24 & 38   & \textbf{450}  &      &       & 42   &     & 554   \\
T2 & MegaFruits-peach     & 28 & 41   & \textbf{482}  &      &       & 34   &     & 585   \\
T2 & PhenoBench       & \textbf{499} & 34   & 28   &      &       & \textbf{171} &     & 732   \\
T2 & PlantSeg         & 75  & \textbf{944}  & 73   &      &       & 55   &     & 1{,}147 \\
\midrule
\multicolumn{2}{@{}l}{\textbf{T2 total}} & 817 & 1{,}076 & 1{,}137 & --- & --- & 515 & --- & \textbf{3{,}545} \\
\bottomrule
\end{tabular}

\vspace{0.4em}
\begin{minipage}{\textwidth}
\footnotesize
\raggedright
T2 columns for pest, tree, and wheat head are empty by construction: the AgroVG-T2 global target family space is $\{$crop, weed, fruit, disease$\}$ (Appendix~\ref{app:query_generation}). Bold cells indicate the source's primary family or families.
\end{minipage}
\end{table}

\paragraph{Per-program-type query counts.}
Table~\ref{tab:program_type_counts} reports the realized query count per program type. T1 instantiates $18$ program types covering single-target uniqueness and rank patterns, multi-target family-all and top-$K$ patterns at both family and fine-class granularity, side-region selection, family-relation queries, and target-absent negatives. T2 instantiates a more restricted library of $12$ program types focused on family-level templates: rank-based queries on individual instances within a fine-grained class often have ambiguous mask-level semantics and were excluded during template design. T2 additionally introduces \texttt{single\_rank\_smallest}, which is absent from T1 because the smallest-by-bbox-area instance is often visually ambiguous, while a smallest-mask instance carries pixel-level evidence that anchors the reference unambiguously.

\begin{table}[t]
\centering
\caption{
\textbf{Per-program-type query counts.}
For each task we list the realized number of queries emitted by each program type, broken down into dev and test splits. Program types are grouped by query type (single, multi, empty). The total counts add up to the per-task headline counts ($6{,}526$ for T1 and $3{,}545$ for T2).
}
\label{tab:program_type_counts}

\setlength{\tabcolsep}{4pt}
\renewcommand{\arraystretch}{1.10}

\begin{tabular}{@{}l l l r r r@{}}
\toprule
\textbf{Task} & \textbf{Query type} & \textbf{Program type} & \textbf{Dev} & \textbf{Test} & \textbf{Total} \\
\midrule
T1 & single & \texttt{single\_family\_unique}      & 212 & 43  & 255 \\
T1 & single & \texttt{single\_fine\_unique}        & 202 & 53  & 255 \\
T1 & single & \texttt{single\_rank\_leftmost}      & 382 & 73  & 455 \\
T1 & single & \texttt{single\_rank\_rightmost}     & 338 & 75  & 413 \\
T1 & single & \texttt{single\_rank\_topmost}       & 371 & 86  & 457 \\
T1 & single & \texttt{single\_rank\_bottommost}    & 360 & 90  & 450 \\
T1 & single & \texttt{single\_rank\_largest}       & 685 & 196 & 881 \\
T1 & single & \texttt{single\_relation\_family}    & 24  & 5   & 29  \\
T1 & multi  & \texttt{multi\_family\_all}              & 402 & 96  & 498 \\
T1 & multi  & \texttt{multi\_family\_side\_left}       & 63  & 15  & 78  \\
T1 & multi  & \texttt{multi\_family\_side\_right}      & 75  & 15  & 90  \\
T1 & multi  & \texttt{multi\_family\_topk\_leftmost}   & 664 & 172 & 836 \\
T1 & multi  & \texttt{multi\_family\_topk\_largest}    & 536 & 132 & 668 \\
T1 & multi  & \texttt{multi\_fine\_all}                & 58  & 18  & 76  \\
T1 & multi  & \texttt{multi\_fine\_side\_right}        & 1   & 0   & 1   \\
T1 & multi  & \texttt{multi\_fine\_topk\_leftmost}     & 12  & 4   & 16  \\
T1 & multi  & \texttt{multi\_fine\_topk\_largest}      & 55  & 16  & 71  \\
T1 & empty  & \texttt{empty\_family\_negative}     & 798 & 199 & 997 \\
\midrule
\multicolumn{3}{@{}l}{\textbf{T1 subtotal}}                 & \textbf{5{,}238} & \textbf{1{,}288} & \textbf{6{,}526} \\
\midrule
T2 & single & \texttt{single\_family\_unique}      & 85  & 20  & 105 \\
T2 & single & \texttt{single\_fine\_unique}        & 81  & 32  & 113 \\
T2 & single & \texttt{single\_rank\_leftmost}      & 204 & 56  & 260 \\
T2 & single & \texttt{single\_rank\_rightmost}     & 211 & 55  & 266 \\
T2 & single & \texttt{single\_rank\_topmost}       & 217 & 40  & 257 \\
T2 & single & \texttt{single\_rank\_bottommost}    & 231 & 48  & 279 \\
T2 & single & \texttt{single\_rank\_largest}       & 217 & 45  & 262 \\
T2 & single & \texttt{single\_rank\_smallest}      & 171 & 59  & 230 \\
T2 & multi  & \texttt{multi\_family\_all}              & 358 & 90  & 448 \\
T2 & multi  & \texttt{multi\_family\_topk\_leftmost}   & 306 & 71  & 377 \\
T2 & multi  & \texttt{multi\_family\_topk\_largest}    & 329 & 87  & 416 \\
T2 & empty  & \texttt{empty\_family\_absent}       & 426 & 106 & 532 \\
\midrule
\multicolumn{3}{@{}l}{\textbf{T2 subtotal}}                 & \textbf{2{,}836} & \textbf{709}  & \textbf{3{,}545} \\
\bottomrule
\end{tabular}
\end{table}

\section{Evaluation Setup and Reproducibility}
\label{app:evaluation_setup}

This section provides reproducibility-oriented details that complement the
evaluation protocol described in Sec.~\ref{sec:evaluation}: full metric
definitions and edge-case conventions (Sec.~\ref{app:metrics}), model
interfaces, prompts, and output parsing rules (Sec.~\ref{app:models_prompts}),
and the compute environment and released artifacts that support score
recomputation (Sec.~\ref{app:compute}).

\subsection{Metric Definitions and Edge Cases}
\label{app:metrics}

Throughout this subsection, $\mathcal{Q}^{+}$ denotes positive queries 
whose ground-truth target set is non-empty, and $\mathcal{Q}^{0}$ 
denotes target-absent queries whose correct output is empty. For T2, 
mask-quality metrics are reported on $\mathcal{Q}^{+}$ and abstention 
metrics on $\mathcal{Q}^{0}$ unless explicitly labeled as all-query 
diagnostics.

\paragraph{T1 box-set matching.}
For a T1 query $q$, let $\mathcal{B}^{*}_q$ be the ground-truth box set and
$\hat{\mathcal{B}}_q$ be the predicted box set after output normalization. At
an IoU threshold $\tau \in \{0.50,0.75\}$, we form a thresholded bipartite
graph between ground-truth and predicted boxes:
\begin{equation}
  E_{\tau}(q)=
\{(b_i^*,\hat b_j): \mathrm{IoU}(b_i^*,\hat b_j)\geq \tau,\ 
b_i^*\in \mathcal{B}^{*}_q,\ \hat b_j\in \hat{\mathcal{B}}_q\}
\end{equation}
We then compute a maximum-cardinality matching
$M_{\tau}(q)\subseteq E_{\tau}(q)$. This objective counts how many instances
are correctly localized at threshold $\tau$, rather than maximizing the sum
of IoU values among matched pairs; the choice reflects AgroVG's set-prediction
formulation, where each ground-truth instance should be claimed by at most one
prediction. Matched pairs are counted as true positives, unmatched predictions
as false positives, and unmatched ground-truth boxes as false negatives:
\begin{equation}
  \mathrm{TP}_{\tau}(q)=|M_{\tau}(q)|,\quad
\mathrm{FP}_{\tau}(q)=|\hat{\mathcal{B}}_q|-|M_{\tau}(q)|,\quad
\mathrm{FN}_{\tau}(q)=|\mathcal{B}^{*}_q|-|M_{\tau}(q)|
\end{equation}
The query-level Set-$F_1$ is the harmonic mean of the corresponding precision
and recall:
\begin{equation}
  F_{1,\tau}(q)=
\frac{2\mathrm{TP}_{\tau}(q)}
{2\mathrm{TP}_{\tau}(q)+\mathrm{FP}_{\tau}(q)+\mathrm{FN}_{\tau}(q)}
\end{equation}
Two boundary conventions handle empty-set cases consistently with the
abstention semantics: $F_{1,\tau}(q)=1$ when both $\mathcal{B}^{*}_q$ and
$\hat{\mathcal{B}}_q$ are empty, and $F_{1,\tau}(q)=0$ when
$\mathcal{B}^{*}_q$ is empty but $\hat{\mathcal{B}}_q$ is not empty.

\paragraph{T1 aggregation, single-target accuracy, and abstention.}
For any query subset $\mathcal{S}\subseteq\mathcal{Q}$, macro Set-$F_1$ averages
query-level scores, while micro Set-$F_1$ first sums true positives, false
positives, and false negatives over $\mathcal{S}$ and then computes a single
$F_1$:

\begin{equation}
  \mathrm{Macro}\text{-}F_{1,\tau}(\mathcal{S})
=
\frac{1}{|\mathcal{S}|}
\sum_{q\in\mathcal{S}}F_{1,\tau}(q)
\end{equation}

\begin{equation}
  \mathrm{Micro}\text{-}F_{1,\tau}(\mathcal{S})
=
\frac{2\sum_{q\in\mathcal{S}} \mathrm{TP}_{\tau}(q)}
{2\sum_{q\in\mathcal{S}} \mathrm{TP}_{\tau}(q)
+\sum_{q\in\mathcal{S}} \mathrm{FP}_{\tau}(q)
+\sum_{q\in\mathcal{S}} \mathrm{FN}_{\tau}(q)}
\end{equation}

Macro Set-$F_1$ weights each query equally and is reported as the headline T1
metric. Micro Set-$F_1$ is computed by the released evaluator and included in
the released summary files as a complementary box-level diagnostic, where
high-cardinality queries contribute more strongly through their instance counts.

For single-target queries, we additionally report a strict accuracy that
penalizes both cardinality errors and localization errors. A single-target
query is counted as correct only when the model predicts exactly one valid
box and that box matches the unique ground-truth box at threshold $\tau$:

\begin{equation}
  \mathrm{S\text{-}Acc}_{\tau}
=
\frac{1}{|\mathcal{Q}_{\mathrm{single}}|}
\sum_{q\in\mathcal{Q}_{\mathrm{single}}}
\mathbb{1}\!\left[
|\hat{\mathcal{B}}_q|=1
\ \wedge\
\max_{b^*\in \mathcal{B}^{*}_q,\hat b\in\hat{\mathcal{B}}_q}
\mathrm{IoU}(b^*,\hat b)\geq \tau
\right]
\end{equation}
For target-absent queries, the correct prediction is the empty set. Empty
Accuracy is the fraction of target-absent queries with $|\hat{\mathcal{B}}_q|=0$:

\begin{equation}
  \mathrm{E\text{-}Acc}
=
\frac{1}{|\mathcal{Q}^{0}|}
\sum_{q\in\mathcal{Q}^{0}}
\mathbb{1}\!\left[|\hat{\mathcal{B}}_q|=0\right]
\end{equation}

Empty Accuracy is reported separately rather than aggregated with positive-query
localization diagnostics, because it measures existence-aware abstention rather
than box-set overlap on non-empty target sets.

\paragraph{T2 query-level mask construction.}
For T2, evaluation operates on query-level binary masks rather than per-instance
mask matching. This design reflects the heterogeneity of segmentation predictors
evaluated in this paper, including per-instance segmentation models, query-level
promptable text-to-mask models, and two-stage box-to-mask pipelines. Let
$L_q\in\mathbb{Z}^{H\times W}$ be the image-level instance map and let
$\mathcal{I}_q$ be the set of target instance IDs associated with query $q$.
The ground-truth query mask is the union of all target instances:

\begin{equation}
  M_q^*(i,j)=\mathbb{1}[L_q(i,j)\in\mathcal{I}_q]
\end{equation}

For target-absent queries, $\mathcal{I}_q=\emptyset$ and $M_q^*$ is the all-zero
mask. The predicted mask $\hat M_q$ is binarized using the model's documented
foreground convention and resized to the ground-truth resolution by
nearest-neighbor interpolation when needed.

\paragraph{T2 mask-quality metrics on positive queries.}
For each query, intersection $I_q$, union $U_q$, ground-truth area $A_q^*$,
and predicted area $\hat A_q$ are computed pixel-wise:
\begin{equation}
  I_q=\sum_{i,j} M_q^*(i,j)\hat M_q(i,j),
\quad
U_q=\sum_{i,j}\mathbb{1}[M_q^*(i,j)\vee \hat M_q(i,j)]
\end{equation}

\begin{equation}
  A_q^*=\sum_{i,j}M_q^*(i,j),
\qquad
\hat A_q=\sum_{i,j}\hat M_q(i,j)
\end{equation}

Per-query IoU and Dice are defined as

\begin{equation}
  \mathrm{IoU}_q=
\begin{cases}
I_q/U_q, & U_q>0,\\
1, & U_q=0,
\end{cases}
\qquad
\mathrm{Dice}_q=
\begin{cases}
2I_q/(A_q^*+\hat A_q), & A_q^*+\hat A_q>0,\\
1, & A_q^*+\hat A_q=0.
\end{cases}
\end{equation}


The value $1$ in the empty-empty case corresponds to correct abstention and is
used only for explicitly labeled all-query diagnostic summaries. The main T2
mask-quality metrics are computed on positive queries only:

\begin{equation}
\mathrm{mIoU}^{+}
=
\frac{1}{|\mathcal{Q}^{+}|}\sum_{q\in\mathcal{Q}^{+}}\mathrm{IoU}_q,
\qquad
\mathrm{mDice}^{+}
=
\frac{1}{|\mathcal{Q}^{+}|}\sum_{q\in\mathcal{Q}^{+}}\mathrm{Dice}_q
\end{equation}



\begin{equation}
  \mathrm{cIoU}^{+}
=
\frac{\sum_{q\in\mathcal{Q}^{+}} I_q}
{\sum_{q\in\mathcal{Q}^{+}} U_q},
\qquad
\mathrm{cDice}^{+}
=
\frac{2\sum_{q\in\mathcal{Q}^{+}} I_q}
{\sum_{q\in\mathcal{Q}^{+}} A_q^*+\sum_{q\in\mathcal{Q}^{+}}\hat A_q}
\end{equation}

Thresholded mask success is reported as

\begin{equation}
  \mathrm{IoU@}\tau^{+}
=
\frac{1}{|\mathcal{Q}^{+}|}
\sum_{q\in\mathcal{Q}^{+}}
\mathbb{1}[\mathrm{IoU}_q\geq \tau],
\qquad
\tau\in\{0.50,0.75\}
\end{equation}

The query-averaged metrics treat each query equally and reflect typical query
difficulty; the cumulative metrics aggregate intersections, unions, and areas
before taking the ratio and are therefore more sensitive to large masks. We use
thresholded success rates rather than AP-style metrics because the evaluated
models either do not expose confidence scores or expose scores with incompatible
calibration.

\paragraph{T2 target-absent metrics.}
For target-absent queries, evaluation measures abstention rather than foreground
overlap. With empty-area tolerance $\delta_e=0$, Empty Accuracy is the fraction
of target-absent queries with $\hat A_q\leq\delta_e$:

\begin{equation}
  \mathrm{E\text{-}Acc}
=
\frac{1}{|\mathcal{Q}^{0}|}
\sum_{q\in\mathcal{Q}^{0}}
\mathbb{1}[\hat A_q\leq \delta_e]
\end{equation}

and the corresponding empty false-positive rate is

\begin{equation}
  \mathrm{Empty\ FPR}=1-\mathrm{E\text{-}Acc}
\end{equation}

As in T1, target-absent behavior is reported as a stand-alone metric so that
pixel-level overlap on positive queries and existence-aware abstention can be
analyzed independently.

\paragraph{Family-macro reporting.}
Family-macro scores are computed by first evaluating the metric within each
target family and then averaging over families with at least one valid query:

\begin{equation}
  \mathrm{FamMacro}(m)
=
\frac{1}{|\mathcal{F}_{\mathrm{eval}}|}
\sum_{f\in\mathcal{F}_{\mathrm{eval}}}
m(\mathcal{Q}_f)
\end{equation}

This prevents large sources or frequent target families from dominating the
aggregate score and makes cross-family differences visible in the diagnostic
breakdowns.

\paragraph{Malformed and missing outputs.}
All predictions pass through the official AgroVG output normalizer before
scoring (Sec.~\ref{app:models_prompts}). Missing or unparsable T1 outputs are
treated as empty box sets, while missing or unparsable T2 outputs are treated
as all-zero masks; degenerate boxes, invalid coordinates, duplicate boxes,
missing masks, and mask-shape mismatches are handled deterministically by the
released evaluator. These conventions ensure that every released query
contributes exactly once to the reported metrics, and that output-format
failures count against the model's score rather than being silently dropped.

\subsection{Evaluated Models, Prompts, and Parsing}
\label{app:models_prompts}

This subsection details the model interfaces, prompts, and output parsing rules
used to produce the predictions evaluated in Sec.~\ref{sec:experiments}. The
26 task-specific configurations are summarized in
Table~\ref{tab:evaluated_model_registry}. They span closed-source MLLMs,
open-source VLMs, and specialized grounding systems evaluated through their
native output interfaces or two-stage grounding-to-mask pipelines. Closed-source
MLLMs and open-source VLMs are evaluated on T1 box grounding, while specialized
systems cover T1 detectors, T2 segmentation MLLMs, T2 promptable text-to-mask
models, and T2 two-stage pipelines. Florence-2-Large and OWLv2-Large appear in
multiple task-specific configurations because they are evaluated through
different output interfaces. We use a fixed interface-level prompting protocol
across all configurations and select detector thresholds and prompt formats only
on the development split.

\paragraph{Prompting protocol.}
Prompts are specified at the interface level rather than tuned per model. For
instruction-following MLLMs and VLMs without a documented native grounding
syntax, T1 queries use the following prompt:
\begin{small}
\begin{verbatim}
You are solving an agricultural visual grounding task.
Given an image and a natural-language query, return all bounding boxes
for objects that match the query.

Query: {query}

Return JSON only:
{"boxes": [[xmin, ymin, xmax, ymax], ...]}

Coordinates must be normalized to [0,1] and use xyxy order.
If no object matches the query, return {"boxes": []}.
Include every matching object and only matching objects.
\end{verbatim}
\end{small}
The prompt specifies the output container, coordinate convention, target-absent
behavior, and cardinality rule. Models with documented native grounding syntax
use their interface-specific formats: DeepSeek-VL2 uses
\texttt{<|ref|>\{query\}<|/ref|>}; Florence-2-Large uses
\texttt{<OPEN\_VOCABULARY\_DETECTION>\{query\}} for T1 and
\texttt{<REFERRING\_EXPRESSION\_SEGMENTATION>\{query\}} for T2; T1 grounding
detectors such as OWLv2-Large and Grounding DINO 1.5 receive the raw AgroVG
query text without an instruction-style preamble. For T2, promptable and
two-stage mask systems receive the raw segmentation query directly, while
segmentation MLLMs that require an explicit segmentation trigger append the
suffix ``\texttt{Please output segmentation mask.}'' to the query. The same
prompt family is used for all queries within a model configuration; prompts are
not retuned on the test split.

\paragraph{T1 output normalization.}
The T1 parser converts heterogeneous model outputs into a canonical list of
absolute-pixel \texttt{xyxy} boxes. Structured JSON outputs are parsed directly
when possible. For nested JSON, tagged box formats, or valid boxes embedded in
free-form text, the parser applies documented recovery rules before conversion
to the AgroVG box format. Coordinate conventions are normalized to absolute
pixels, including pixel coordinates, $[0,1]$ normalized coordinates, and
$[0,1000]$ coordinate formats used by some VLMs. Florence-2 outputs are
processed by its official post-processor before conversion; DeepSeek-VL2 tagged
boxes are converted from the model's normalized coordinate system to image
pixels. All boxes are clipped to image bounds, and predictions with non-positive
area, degenerate full-image extent, or exact duplication are removed before
scoring.

\paragraph{T2 output normalization.}
The T2 parser converts each model output into one query-level binary mask saved
as a grayscale PNG. Direct mask outputs are binarized and unioned via
pixel-wise logical OR when multiple masks are returned. Polygon outputs are
rasterized onto the original image grid and unioned. Segmentation-token models
are decoded through their native mask decoders, and all \texttt{[SEG]}-derived
masks are unioned. For two-stage pipelines, masks produced by SAM from each
detector box are unioned into a single query-level prediction. If a model
produces no usable mask, the prediction is represented as an all-zero mask saved
at the canonical path, ensuring deterministic representation at evaluation time.
Each prediction record stores the query ID and the mask path, allowing the
official evaluator to recompute all T2 metrics without rerunning inference.

\paragraph{Detector thresholds and parser warnings.}
Detector thresholds, when required, are selected on the development split and
held fixed for the test split. The official evaluator emits per-run warning
files documenting missing predictions, malformed boxes, missing mask paths,
mask-shape mismatches, and other normalization events. These warning logs are
released alongside the predictions so that output-format failures can be audited
separately from model-level grounding errors.

\begin{table}[!t]
\centering
\setlength{\abovecaptionskip}{2pt}
\setlength{\belowcaptionskip}{2pt}
\caption{
\textbf{Evaluated model registry.}
We report the 26 task-specific configurations evaluated on AgroVG, including
API identifiers or public checkpoints, task assignment, and output interface.
``Native'' indicates that the model directly emits the evaluated output, while
``Two-stage'' indicates detector-based grounding followed by SAM mask generation.
}
\label{tab:evaluated_model_registry}

\fontsize{6.3pt}{6.9pt}\selectfont
\setlength{\tabcolsep}{2.5pt}
\renewcommand{\arraystretch}{0.88}

\resizebox{0.96\textwidth}{!}{%
\begin{tabular}{@{}l c c l l@{}}
\toprule
\textbf{Model} & \textbf{Params.} & \textbf{Task} & \textbf{API identifier / checkpoint} & \textbf{Interface} \\
\midrule

\rowcolor{gray!15}
\multicolumn{5}{@{}l}{\textbf{\textit{Closed-source MLLMs}}} \\
GPT-4o
& -- & T1
& \texttt{gpt-4o-2024-08-06}
& OpenAI API, box output \\

GPT-5.4
& -- & T1
& \texttt{gpt-5.4}
& OpenAI API, box output \\

Gemini 2.5 Pro
& -- & T1
& \texttt{gemini-2.5-pro}
& Google API, box output \\

Claude Sonnet 4.6
& -- & T1
& \texttt{claude-sonnet-4-6}
& Anthropic API, box output \\

\midrule

\rowcolor{gray!15}
\multicolumn{5}{@{}l}{\textbf{\textit{Open-source VLMs}}} \\
DeepSeek-VL2-tiny
& $\sim$1B & T1
& \texttt{deepseek-ai/deepseek-vl2-tiny}
& Native / prompted box output \\

DeepSeek-VL2-small
& $\sim$2.8B & T1
& \texttt{deepseek-ai/deepseek-vl2-small}
& Native / prompted box output \\

InternVL3.5-8B
& 8.53B & T1
& \texttt{OpenGVLab/InternVL3\_5-8B}
& Prompted box output \\

InternVL3.5-38B
& 38.4B & T1
& \texttt{OpenGVLab/InternVL3\_5-38B}
& Prompted box output \\

Qwen3-VL-8B
& 9B & T1
& \texttt{Qwen/Qwen3-VL-8B-Instruct}
& Prompted box output \\

Qwen3-VL-32B
& 33B & T1
& \texttt{Qwen/Qwen3-VL-32B-Instruct}
& Prompted box output \\

\midrule

\rowcolor{gray!15}
\multicolumn{5}{@{}l}{\textbf{\textit{Specialized T1 grounding detectors}}} \\
OWLv2-Large
& 0.4B & T1
& \texttt{google/owlv2-large-patch14-ensemble}
& Native box output \\

Florence-2-Large
& 0.77B & T1
& \texttt{microsoft/Florence-2-large}
& Native box output \\

Grounding DINO 1.5
& -- & T1
& \texttt{IDEA-Research/Grounding-DINO-1.5-API}
& Native box output \\

\midrule

\rowcolor{gray!15}
\multicolumn{5}{@{}l}{\textbf{\textit{T2 segmentation MLLMs / reasoning-segmentation models}}} \\
LISA-7B
& 7B & T2
& \texttt{xinlai/LISA-7B-v1}
& Native mask output \\

LISA-13B
& 13B & T2
& \texttt{xinlai/LISA-13B-llama2-v1}
& Native mask output \\

GLaMM
& 7B & T2
& \texttt{MBZUAI/groundingLMM-FullScope}
& Native mask output \\

PixelLM-7B
& 7B & T2
& \texttt{MaverickRen/PixelLM-7B}
& Native mask output \\

PixelLM-13B
& 13B & T2
& \texttt{MaverickRen/PixelLM-13B}
& Native mask output \\

GSVA (force-seg)
& 13B & T2
& \texttt{LeapLabTHU/GSVA-13B-ft}
& Native mask output \\

PSALM
& 2B & T2
& \texttt{EnmingZhang/PSALM}
& Native mask output \\

\midrule

\rowcolor{gray!15}
\multicolumn{5}{@{}l}{\textbf{\textit{T2 promptable / universal text-to-mask models}}} \\
X-Decoder
& -- & T2
& \texttt{microsoft/X-Decoder}
& Native mask output \\

SEEM
& -- & T2
& \texttt{xdecoder/SEEM}
& Native mask output \\

Florence-2-Large
& 0.77B & T2
& \texttt{microsoft/Florence-2-large}
& Native mask output \\

SAM~3
& 0.9B & T2
& \texttt{facebook/sam3}
& Native mask output \\

\midrule

\rowcolor{gray!15}
\multicolumn{5}{@{}l}{\textbf{\textit{T2 two-stage box-to-mask pipelines}}} \\
Grounded-SAM
& Comp. & T2
& GroundingDINO-T + \texttt{facebook/sam-vit-huge}
& Two-stage \\

OWLv2-Large + SAM
& Comp. & T2
& OWLv2-Large + \texttt{facebook/sam-vit-huge}
& Two-stage \\

\bottomrule
\end{tabular}%
}

\vspace{0.15em}
\begin{minipage}{0.96\textwidth}
\scriptsize
\raggedright
For API-based models, released run metadata records the provider identifier,
query date, prompt family, decoding parameters, and raw response before parsing.
For local models, run manifests record checkpoint, precision, decoding settings,
thresholds where applicable, and parser version. ``Comp.'' denotes a composite
pipeline rather than a single parameterized model.
\end{minipage}
\end{table}

\subsection{Compute and Reproducibility Artifacts}
\label{app:compute}

AgroVG evaluates existing models without training or fine-tuning, so the
benchmark's compute footprint is dominated by inference. This subsection
documents the compute environment, the released artifacts, and the two
reproducibility modes that the release supports.

\paragraph{Compute environment.}
Open-source VLMs and specialized grounding systems are evaluated locally on
NVIDIA A100 80GB GPUs using each model's official loader, checkpoint, and
preprocessing pipeline. Closed-source MLLMs are queried through their official
APIs. For every run, we record a manifest containing the checkpoint identifier
or provider model ID, query date for API runs, code revision when available,
precision and decoding settings, detector thresholds, prompt family, parser
version, hardware configuration, and the raw output before parsing. We use
deterministic or greedy decoding whenever supported, and detector thresholds
selected on the development split are fixed before test-set evaluation. The
local evaluation campaign for open-source and specialized configurations
required approximately 320 A100 GPU-hours in total, covering inference, output
normalization, and metric computation; closed-source API compute is borne by
the providers and is not observable from the client.

\paragraph{Released evaluation artifacts.}
The AgroVG release contains all non-provider artifacts needed to audit and
recompute the reported metrics, organized into three groups. \emph{Data:}
fixed dev/test split files, normalized image and instance metadata, and query
JSONL files. \emph{Protocol:} prompt templates, parsing scripts, and the
official evaluation scripts. \emph{Predictions:} per-model prediction JSONLs,
saved T2 mask predictions, per-query result files, aggregate
\texttt{summary.json} files, and parser-warning logs. Every released query maps
to exactly one normalized prediction before scoring, so the official evaluator
reproduces the paper tables from the released predictions without rerunning any
model inference.

\paragraph{Reproducibility modes.}
The released artifacts support two reproducibility modes. \emph{Score
recomputation:} users run the official evaluator on the released predictions
and recover the aggregate, per-source, per-family, per-query-type, and
per-program-type metrics reported in the paper; this mode requires no GPU
inference. \emph{Full inference reproduction:} users obtain the original source
images under their respective licenses, load the listed checkpoints or query
the listed APIs, apply the released prompt and parsing protocol, and rerun the
evaluator. Full reproduction is intended to be deterministic for locally
evaluated models under the recorded decoding settings, software versions, and
parsing rules; minor numerical differences may still arise from hardware or
library changes. For closed-source APIs, exact reproduction may differ because
provider-side model serving is outside the benchmark's control. The released raw
responses, parsed predictions, model identifiers, and query dates define the
auditable record for the reported API scores and allow the published tables to
be recomputed even if a provider later changes the model behind a public
endpoint.

\section{Additional Results and Diagnostics}
\label{app:additional_results}

\subsection{T1 Diagnostic Results}
\label{app:extended_t1}

This subsection expands the AgroVG-T1 results in 
Table~\ref{tab:t1_main_results}. Unless otherwise stated, we report 
macro Set-$F_1$ at IoU $\tau=0.50$, matching the headline T1 diagnostic 
used for source and program-level analysis. All numbers are computed from the 
released prediction files and \texttt{summary.json} outputs; the rightmost 
``All'' column of the per-source table reproduces the corresponding overall 
score in Table~\ref{tab:t1_main_results} as a consistency check.

\paragraph{Per-source robustness.}
Table~\ref{tab:t1_per_source_50} reports source-level macro Set-$F_1$ at IoU
$0.50$. Sources differ in viewpoint, target scale, instance density, and visual
clutter, and the breakdown makes these source-specific transfer behaviors
explicit. Models with similar aggregate scores can have different per-source
profiles, indicating that the global T1 score hides meaningful variation across
agricultural subdomains. Strict localization is summarized by the aggregate
All@.75 column in Table~\ref{tab:t1_main_results}, while the released summaries
include the full thresholded breakdowns for further analysis.

\paragraph{Per-query-type decomposition.}
Table~\ref{tab:t1_per_query_type} decomposes performance into single-target,
multi-target, and target-absent regimes, exposing complementary failure modes.
Single-target Set-$F_1$ measures referent binding, while strict accuracy
(S-Acc) additionally penalizes cardinality errors. Multi-target Set-$F_1$
captures set completeness and over-prediction. Empty Accuracy measures
abstention when the referred target family is absent. The three regimes are
not strongly correlated across models: DeepSeek-VL2-tiny achieves $97.5\%$
Empty Accuracy but only $10.3\%$ single-target $F_1$, while
Florence-2-Large achieves $26.6\%$ single-target $F_1$ but $0\%$ Empty
Accuracy. This decoupling supports reporting all three regimes jointly rather
than collapsing them into a single aggregate.

\paragraph{Per-program-type sensitivity.}
Table~\ref{tab:t1_per_program_bucket} aggregates the 18 T1 program types from
Appendix~\ref{app:query_generation} into eight diagnostic buckets. Single-unique
queries primarily test family or class identification under minimal cardinality
ambiguity. Spatial-rank and size-rank queries require ordering among visually
similar instances. Multi-target all and top-$K$ queries test set completion.
Side-selection and family-relation queries test spatially constrained set
selection. The program-level view separates open-vocabulary detection ability
from referring-expression reasoning: nearly all models score substantially
higher on Single-unique than on rank-based or relation buckets, even when their
aggregate Set-$F_1$ is similar.

\begin{table}[t]
\centering
\caption{
\textbf{T1 per-source macro Set-$F_1$ at IoU $\tau = 0.50$.}
Each cell is the macro-averaged Set-$F_1$ over the model's per-query results
within the corresponding source. The rightmost column reproduces the overall
T1 score from Table~\ref{tab:t1_main_results} as a consistency check. All
numbers are percentages.
}
\label{tab:t1_per_source_50}

\scriptsize
\setlength{\tabcolsep}{3pt}
\renewcommand{\arraystretch}{1.10}

\begin{tabular}{@{}l ccccccccc c@{}}
\toprule
\textbf{Model} & \textbf{ACFR} & \textbf{CW} & \textbf{Dumort.} & \textbf{GWHD} & \textbf{IP102} & \textbf{Minne.} & \textbf{OAM-TCD} & \textbf{Pheno.} & \textbf{PlantSeg} & \textbf{All} \\
\midrule
\multicolumn{11}{@{}l}{\emph{Closed-source MLLMs}} \\
GPT-4o                   & 3.1 & 20.9 & 15.8 & 17.3 & 22.6 & 15.3 & 14.3 & 8.5 & 11.1 & 15.0 \\
GPT-5.4                  & 8.8 & 37.6 & 19.5 & 16.0 & 62.7 & 16.0 & 14.3 & 28.3 & 13.9 & 27.4 \\
Gemini 2.5 Pro           & 31.0 & 18.4 & 11.5 & 7.4 & 28.0 & 13.5 & 4.8 & 35.5 & 10.3 & 18.5 \\
Claude Sonnet 4.6        & 6.7 & 20.9 & 13.8 & 8.5 & 48.6 & 11.9 & 15.9 & 25.5 & 10.5 & 20.0 \\
\midrule
\multicolumn{11}{@{}l}{\emph{Open-source VLMs}} \\
DeepSeek-VL2-tiny        & 5.4 & 23.3 & 17.0 & 17.8 & 37.7 & 41.0 & 14.3 & 15.7 & 14.7 & 22.3 \\
DeepSeek-VL2-small       & 18.2 & 45.4 & 26.2 & 10.4 & 70.5 & 31.5 & 21.6 & 29.2 & 14.4 & 31.9 \\
InternVL3.5-8B           & 62.3 & 38.4 & 34.8 & 18.3 & 71.4 & 39.9 & 20.2 & 38.2 & 17.2 & 39.6 \\
InternVL3.5-38B          & 61.7 & 31.8 & 44.9 & 22.2 & 73.4 & 24.8 & 26.2 & 39.6 & 18.9 & 39.8 \\
Qwen3-VL-8B              & 36.0 & 21.4 & 29.8 & 23.8 & 63.3 & 29.1 & 14.3 & 33.4 & 16.8 & 32.7 \\
Qwen3-VL-32B             & 55.7 & 21.2 & 41.2 & 24.5 & 73.8 & 15.3 & 14.3 & 35.2 & 22.1 & 37.5 \\
\midrule
\multicolumn{11}{@{}l}{\emph{Specialized grounding detectors}} \\
OWLv2-Large              & 28.8 & 46.8 & 16.7 & 18.0 & 51.5 & 34.4 & 29.7 & 28.7 & 10.5 & 29.5 \\
Florence-2-Large         & 38.9 & 25.4 & 9.3 & 1.7 & 37.2 & 21.3 & 14.6 & 20.3 & 9.6 & 20.2 \\
Grounding DINO 1.5       & 38.1 & 31.3 & 13.3 & 1.9 & 54.9 & 15.6 & 10.3 & 21.6 & 13.5 & 24.7 \\
\bottomrule
\end{tabular}

\vspace{0.3em}
\begin{minipage}{\textwidth}
\footnotesize
\raggedright
CW = CropAndWeed; Dumort. = Dumortier; Minne. = MinneApple; Pheno. = PhenoBench.
\end{minipage}
\end{table}

\begin{table}[t]
\centering
\caption{
\textbf{T1 per-query-type breakdown at IoU $\tau = 0.50$.}
For single-target and multi-target queries we report macro Set-$F_1$ (\%); for
target-absent queries we report Empty Accuracy (\%). The single-target column
also reports strict accuracy S-Acc (\%) for reference. All numbers are
percentages.
}
\label{tab:t1_per_query_type}

\setlength{\tabcolsep}{4pt}
\renewcommand{\arraystretch}{1.10}

\begin{tabular}{@{}l cc c c@{}}
\toprule
\textbf{Model} & \textbf{Single $F_1$} & \textbf{Single S-Acc} & \textbf{Multi $F_1$} & \textbf{Empty E-Acc} \\
\midrule
GPT-4o                   & 4.5 & 4.5 & 1.5 & 79.9 \\
GPT-5.4                  & 22.9 & 22.9 & 16.5 & 66.8 \\
Gemini 2.5 Pro           & 14.3 & 14.0 & 15.8 & 37.7 \\
Claude Sonnet 4.6        & 16.9 & 16.9 & 11.3 & 49.7 \\
DeepSeek-VL2-tiny        & 10.3 & 10.3 & 6.2 & 97.5 \\
DeepSeek-VL2-small       & 30.8 & 30.0 & 23.4 & 55.3 \\
InternVL3.5-8B           & 36.8 & 34.3 & 35.4 & 58.3 \\
InternVL3.5-38B          & 35.7 & 35.3 & 29.8 & 75.9 \\
Qwen3-VL-8B              & 26.3 & 24.2 & 17.4 & 88.9 \\
Qwen3-VL-32B             & 32.0 & 31.9 & 26.3 & 80.9 \\
OWLv2-Large              & 19.7 & 8.1 & 15.8 & 92.5 \\
Florence-2-Large         & 26.6 & 25.8 & 20.3 & 0.0 \\
Grounding DINO 1.5       & 28.8 & 14.2 & 29.1 & 1.5 \\
\bottomrule
\end{tabular}
\end{table}

\begin{table}[t]
\centering
\caption{
\textbf{T1 per-program-bucket macro Set-$F_1$ at IoU $\tau = 0.50$.}
The 18 T1 program types (Appendix~\ref{app:query_generation}) are aggregated
into 8 program-type buckets. Each cell is the count-weighted average of the
constituent program-type macro $F_1$ values. All numbers are percentages.
}
\label{tab:t1_per_program_bucket}

\small
\setlength{\tabcolsep}{3pt}
\renewcommand{\arraystretch}{1.10}

\begin{tabular}{@{}l cccc cccc@{}}
\toprule
\textbf{Model} & \textbf{S-uniq.} & \textbf{S-rank-sp.} & \textbf{S-rank-size} & \textbf{S-rel.} & \textbf{M-all} & \textbf{M-topk} & \textbf{M-side} & \textbf{Empty} \\
\midrule
GPT-4o                   & 15.6 & 1.2 & 4.6 & 0.0 & 3.8 & 0.8 & 0.0 & 79.9 \\
GPT-5.4                  & 67.7 & 15.4 & 13.8 & 0.0 & 32.7 & 10.8 & 17.6 & 66.8 \\
Gemini 2.5 Pro           & 15.6 & 16.4 & 10.7 & 0.0 & 27.9 & 11.2 & 19.9 & 37.7 \\
Claude Sonnet 4.6        & 59.4 & 8.3 & 10.7 & 0.0 & 24.1 & 7.3 & 6.3 & 49.7 \\
DeepSeek-VL2-tiny        & 6.2 & 12.3 & 9.2 & 0.0 & 3.7 & 5.1 & 26.8 & 97.5 \\
DeepSeek-VL2-small       & 66.9 & 27.7 & 17.5 & 60.0 & 50.7 & 14.3 & 18.7 & 55.3 \\
InternVL3.5-8B           & 77.6 & 36.2 & 17.9 & 40.0 & 55.7 & 28.7 & 30.6 & 58.3 \\
InternVL3.5-38B          & 79.1 & 32.4 & 20.1 & 33.3 & 47.0 & 23.5 & 33.1 & 75.9 \\
Qwen3-VL-8B              & 65.9 & 24.1 & 10.6 & 20.0 & 28.7 & 13.6 & 15.0 & 88.9 \\
Qwen3-VL-32B             & 70.7 & 29.6 & 17.3 & 20.0 & 42.7 & 19.9 & 32.2 & 80.9 \\
OWLv2-Large              & 62.4 & 13.5 & 9.1 & 17.4 & 24.3 & 13.3 & 10.5 & 92.5 \\
Florence-2-Large         & 63.5 & 18.5 & 21.0 & 53.3 & 38.8 & 15.1 & 6.9 & 0.0 \\
Grounding DINO 1.5       & 77.2 & 20.2 & 19.3 & 33.3 & 54.6 & 21.2 & 17.9 & 1.5 \\
\bottomrule
\end{tabular}

\vspace{0.3em}
\begin{minipage}{\textwidth}
\footnotesize
\raggedright
S-uniq. = Single-unique (\texttt{single\_fine\_unique} + \texttt{single\_family\_unique});
S-rank-sp. = Single-rank-spatial (4 spatial directions);
S-rank-size = \texttt{single\_rank\_largest};
S-rel. = \texttt{single\_relation\_family};
M-all = \texttt{multi\_family\_all} + \texttt{multi\_fine\_all};
M-topk = 4 topk templates;
M-side = side-selection templates realized in the test split;
Empty = \texttt{empty\_family\_negative} (Empty Accuracy reported in this column).
\end{minipage}
\end{table}

\paragraph{Summary of T1 diagnostics.}
The three breakdowns support three observations consistent with AgroVG's design
intent. First, source-level performance is not fully predicted by the aggregate
score, indicating meaningful variation across agricultural subdomains. Second,
positive localization, set completeness, and target-absent abstention are
partially independent capabilities, with several models trading one for another.
Third, rank-based and set-selection programs remain substantially harder than
unique-target programs across model groups, indicating that agricultural
grounding requires referring-expression reasoning in addition to
open-vocabulary detection. These observations motivate reporting T1 along
multiple diagnostic axes rather than collapsing it to a single leaderboard
number.

\subsection{T2 Diagnostic Results}
\label{app:extended_t2}
This subsection expands the AgroVG-T2 results in Table~\ref{tab:t2_main_results}. Following Appendix~\ref{app:metrics}, positive-query mask quality and target-absent abstention are reported as separate signals: the main T2 mask scores in the paper use $\mathcal{Q}^{+}$ only, while Empty Accuracy is reported on $\mathcal{Q}^{0}$. We include source- and family-level all-query mIoU only as auxiliary diagnostics of joint segmentation and abstention behavior; the primary T2 interpretation should rely on the positive-query and empty-query breakdowns below.

\paragraph{Per-source diagnostics.}
Table~\ref{tab:t2_per_source_miou} reports source-level all-query mIoU. Because this table includes both positive and target-absent queries, its values reflect two coupled behaviors: how accurately a model segments present targets and how reliably it abstains when the queried target is absent. We keep this table as an auxiliary source-level diagnostic, while positive-query mask quality and empty-query abstention are separated explicitly in Tables~\ref{tab:t2_per_query_type} and~\ref{tab:t2_empty_breakdown}.

\paragraph{Per-target-family and per-query-type diagnostics.}
Table~\ref{tab:t2_per_target_family} reports all-query mIoU by requested target family (crop, weed, fruit, disease), providing an auxiliary view of family-conditioned joint segmentation and abstention behavior. Disease and weed queries tend to be more challenging than fruit queries, reflecting smaller, more irregular masks and visually similar surrounding background. Table~\ref{tab:t2_per_query_type} is the primary regime-level diagnostic: it reports positive-query mIoU and IoU@0.50 on single-target and multi-target queries, and Empty Accuracy on target-absent queries, thereby separating segmentation quality from abstention behavior.

\paragraph{Per-program-type diagnostics.}
Table~\ref{tab:t2_per_program_bucket} aggregates the 12 T2 program types into
six diagnostic buckets. Single-unique and Single-rank-spatial buckets test
referent binding under different forms of cardinality and spatial cueing; the
Single-rank-size bucket additionally exercises the smallest-mask referent through
\texttt{single\_rank\_smallest}. Multi-target buckets test coverage of a
query-level union mask, and the Empty bucket reports abstention. The program-level
view identifies whether errors arise from mask extraction (low S-uniq.), instance
selection (low rank buckets), set coverage (low M-all / M-topk), or
target-existence calibration (low Empty).

\paragraph{Empty-query source breakdown.}
Table~\ref{tab:t2_empty_breakdown} reports Empty Accuracy on the target-absent
subset, broken down by source. Abstention failures are highly model- and
source-dependent. SAM~3 abstains correctly on most target-absent queries across
sources, while LISA-style and PixelLM-style models hallucinate foreground on
nearly every target-absent query. Some promptable systems, such as X-Decoder
and SEEM, abstain reliably on PhenoBench and CropAndWeed but much less reliably
on MegaFruits subsets, suggesting that abstention depends on both query content
and visual context.

\begin{table}[t]
\centering
\caption{
\textbf{T2 per-source all-query mIoU diagnostic.}
Each cell averages per-query IoU over all T2 test queries from the corresponding
source, including positive and target-absent queries under the empty-mask
convention in Appendix~\ref{app:metrics}. The table therefore reflects both
positive-query mask quality and target-absent abstention. All numbers are
percentages.
}
\label{tab:t2_per_source_miou}

\setlength{\tabcolsep}{4pt}
\renewcommand{\arraystretch}{1.10}

\begin{tabular}{@{}l cccccc c@{}}
\toprule
\textbf{Model} & \textbf{Pheno.} & \textbf{CW} & \textbf{Blueb.} & \textbf{Peach} & \textbf{ACFR} & \textbf{PlantSeg} & \textbf{All} \\
\midrule
\multicolumn{8}{@{}l}{\emph{Reasoning-segmentation MLLMs}} \\
LISA-7B                & 32.9 & 32.9 & 28.6 & 31.7 & 18.5 & 17.4 & 26.7 \\
LISA-13B               & 30.1 & 37.4 & 30.6 & 42.8 & 12.6 & 15.8 & 28.1 \\
GLaMM                  & 27.0 & 26.1 & 26.7 & 24.2 & 14.3 & 17.2 & 22.9 \\
PixelLM-7B             & 24.0 & 28.5 & 32.5 & 41.7 & 26.4 & 15.7 & 26.2 \\
PixelLM-13B            & 20.2 & 29.1 & 32.3 & 41.1 & 25.4 & 14.6 & 25.0 \\
GSVA (force-seg)      & 20.9 & 17.3 & 15.6 & 23.6 & 6.3 & 16.7 & 18.4 \\
PSALM                  & 29.4 & 23.5 & 29.2 & 41.1 & 33.6 & 13.0 & 25.3 \\
\midrule
\multicolumn{8}{@{}l}{\emph{Promptable / universal text-to-mask models}} \\
X-Decoder              & 10.5 & 18.6 & 18.4 & 26.7 & 31.1 & 17.8 & 18.3 \\
SEEM                   & 20.1 & 23.3 & 19.1 & 29.5 & 25.8 & 14.6 & 20.3 \\
Florence-2-Large       & 10.5 & 17.4 & 28.5 & 30.3 & 20.5 & 14.3 & 18.9 \\
SAM~3                  & 12.7 & 9.6 & 32.6 & 42.8 & 37.3 & 21.5 & 23.8 \\
\midrule
\multicolumn{8}{@{}l}{\emph{Two-stage box-to-mask pipelines}} \\
Grounded-SAM           & 23.4 & 31.4 & 17.6 & 24.0 & 11.9 & 16.1 & 21.0 \\
OWLv2-Large + SAM      & 36.8 & 9.2 & 16.6 & 18.3 & 11.8 & 13.6 & 19.0 \\
\bottomrule
\end{tabular}

\vspace{0.3em}

\end{table}

\begin{table}[t]
\centering
\caption{
\textbf{T2 per-target-family all-query mIoU diagnostic.}
Each cell averages per-query IoU over queries whose
\texttt{meta.target\_family} matches the column. For target-absent queries, the
column corresponds to the requested absent family stored in the query metadata.
The table jointly reflects family-specific mask quality and abstention behavior.
All numbers are percentages.
}
\label{tab:t2_per_target_family}

\setlength{\tabcolsep}{4pt}
\renewcommand{\arraystretch}{1.10}

\begin{tabular}{@{}l cccc c@{}}
\toprule
\textbf{Model} & \textbf{Crop} & \textbf{Weed} & \textbf{Fruit} & \textbf{Disease} & \textbf{All} \\
\midrule
LISA-7B                & 35.1 & 16.4 & 30.9 & 20.7 & 26.7 \\
LISA-13B               & 35.2 & 19.0 & 33.6 & 21.3 & 28.1 \\
GLaMM                  & 29.4 & 12.3 & 25.5 & 20.5 & 22.9 \\
PixelLM-7B             & 28.9 & 12.8 & 38.3 & 17.4 & 26.2 \\
PixelLM-13B            & 25.6 & 13.1 & 37.9 & 16.1 & 25.0 \\
GSVA (force-seg)      & 23.5 & 7.5 & 19.7 & 18.5 & 18.4 \\
PSALM                  & 33.6 & 9.1 & 37.0 & 14.4 & 25.3 \\
X-Decoder              & 9.2 & 18.6 & 31.7 & 10.1 & 18.3 \\
SEEM                   & 17.9 & 13.1 & 29.7 & 15.2 & 20.3 \\
Florence-2-Large       & 14.0 & 7.8 & 33.7 & 11.9 & 18.9 \\
SAM~3                  & 18.9 & 27.5 & 36.3 & 11.7 & 23.8 \\
Grounded-SAM           & 28.6 & 13.9 & 21.8 & 17.9 & 21.0 \\
OWLv2-Large + SAM      & 29.4 & 23.7 & 16.2 & 11.7 & 19.0 \\
\bottomrule
\end{tabular}
\end{table}

\begin{table}[t]
\centering
\caption{
\textbf{T2 per-query-type breakdown.}
For single-target and multi-target positive queries, we report mIoU and
IoU@0.50. For target-absent queries, we report Empty Accuracy. Unlike the
all-query diagnostic tables, this table keeps positive mask quality and
abstention behavior separate. All numbers are percentages.
}
\label{tab:t2_per_query_type}

\setlength{\tabcolsep}{4pt}
\renewcommand{\arraystretch}{1.10}

\begin{tabular}{@{}l cc cc c@{}}
\toprule
& \multicolumn{2}{c}{\textbf{Single}} & \multicolumn{2}{c}{\textbf{Multi}} & \textbf{Empty} \\
\cmidrule(lr){2-3} \cmidrule(lr){4-5}
\textbf{Model} & \textbf{mIoU} & \textbf{IoU@.5} & \textbf{mIoU} & \textbf{IoU@.5} & \textbf{E-Acc} \\
\midrule
LISA-7B                & 23.2 & 21.4 & 41.8 & 44.4 & 2.8 \\
LISA-13B               & 25.0 & 21.7 & 41.5 & 44.0 & 7.5 \\
GLaMM                  & 25.5 & 22.5 & 27.4 & 21.4 & 3.8 \\
PixelLM-7B             & 28.7 & 28.2 & 33.8 & 29.8 & 0.0 \\
PixelLM-13B            & 26.4 & 25.1 & 32.7 & 29.4 & 1.9 \\
GSVA (force-seg)      & 16.2 & 11.5 & 29.3 & 27.4 & 0.0 \\
PSALM                  & 26.4 & 26.5 & 34.7 & 33.5 & 0.0 \\
X-Decoder              & 11.8 & 11.3 & 18.8 & 17.3 & 38.7 \\
SEEM                   & 14.0 & 13.5 & 25.4 & 25.8 & 29.2 \\
Florence-2-Large       & 21.4 & 18.9 & 14.9 & 6.0 & 19.8 \\
SAM~3                  & 9.3 & 6.2 & 14.9 & 15.3 & 93.4 \\
Grounded-SAM           & 21.4 & 21.7 & 28.9 & 25.4 & 0.9 \\
OWLv2-Large + SAM      & 8.6 & 8.2 & 13.0 & 10.5 & 67.9 \\
\bottomrule
\end{tabular}
\end{table}

\begin{table}[t]
\centering
\caption{
\textbf{T2 per-program-bucket diagnostics.}
The 12 T2 program types are aggregated into six diagnostic buckets. Non-empty
program buckets report count-weighted mIoU over positive queries, while the
Empty bucket reports Empty Accuracy on target-absent queries. All numbers are
percentages.
}
\label{tab:t2_per_program_bucket}

\setlength{\tabcolsep}{4pt}
\renewcommand{\arraystretch}{1.10}

\begin{tabular}{@{}l cccccc@{}}
\toprule
\textbf{Model} & \textbf{S-uniq.} & \textbf{S-rank-sp.} & \textbf{S-rank-size} & \textbf{M-all} & \textbf{M-topk} & \textbf{Empty} \\
\midrule
LISA-7B                & 46.0 & 21.0 & 16.0 & 54.2 & 34.8 & 2.8 \\
LISA-13B               & 45.3 & 23.6 & 17.5 & 50.5 & 36.3 & 7.5 \\
GLaMM                  & 48.7 & 23.3 & 18.2 & 29.5 & 26.2 & 3.8 \\
PixelLM-7B             & 42.1 & 30.1 & 19.4 & 41.1 & 29.6 & 0.0 \\
PixelLM-13B            & 36.8 & 27.8 & 18.8 & 38.6 & 29.4 & 1.9 \\
GSVA (force-seg)      & 31.8 & 13.8 & 12.9 & 29.8 & 29.0 & 0.0 \\
PSALM                  & 38.9 & 26.0 & 20.8 & 46.3 & 28.2 & 0.0 \\
X-Decoder              & 17.4 & 10.1 & 12.2 & 30.3 & 12.3 & 38.7 \\
SEEM                   & 22.4 & 11.1 & 15.3 & 38.3 & 18.1 & 29.2 \\
Florence-2-Large       & 22.4 & 24.7 & 14.7 & 15.4 & 14.6 & 19.8 \\
SAM~3                  & 9.4 & 9.3 & 9.3 & 27.0 & 8.0 & 93.4 \\
Grounded-SAM           & 50.0 & 15.7 & 18.1 & 31.7 & 27.3 & 0.9 \\
OWLv2-Large + SAM      & 12.8 & 8.0 & 7.8 & 11.3 & 13.9 & 67.9 \\
\bottomrule
\end{tabular}

\vspace{0.3em}
\begin{minipage}{\textwidth}
\footnotesize
\raggedright
S-uniq. = Single-unique;
S-rank-sp. = Single-rank-spatial (4 spatial directions);
S-rank-size = \texttt{single\_rank\_largest} + \texttt{single\_rank\_smallest};
M-all = \texttt{multi\_family\_all};
M-topk = 2 topk templates;
Empty = Empty Accuracy on \texttt{empty\_family\_absent}.
\end{minipage}
\end{table}

\begin{table}[t]
\centering
\caption{
\textbf{T2 per-source Empty Accuracy on target-absent queries.}
Each cell is the fraction of target-absent queries within the source for which
the model emits no foreground mask, i.e., $\hat{A}_q \leq \delta_e = 0$.
The Overall column reports Empty Accuracy across all target-absent T2 queries.
All numbers are percentages.
}
\label{tab:t2_empty_breakdown}

\setlength{\tabcolsep}{4pt}
\renewcommand{\arraystretch}{1.10}

\begin{tabular}{@{}l cccccc c@{}}
\toprule
\textbf{Model} & \textbf{Pheno.} & \textbf{CW} & \textbf{Blueb.} & \textbf{Peach} & \textbf{ACFR} & \textbf{PlantSeg} & \textbf{All} \\
\midrule
LISA-7B                & 21.4 & 0.0 & 0.0 & 0.0 & 0.0 & 0.0 & 2.8 \\
LISA-13B               & 0.0 & 0.0 & 10.5 & 33.3 & 0.0 & 0.0 & 7.5 \\
GLaMM                  & 14.3 & 0.0 & 0.0 & 11.1 & 0.0 & 0.0 & 3.8 \\
PixelLM-7B             & 0.0 & 0.0 & 0.0 & 0.0 & 0.0 & 0.0 & 0.0 \\
PixelLM-13B            & 0.0 & 0.0 & 0.0 & 5.6 & 0.0 & 2.1 & 1.9 \\
GSVA (force-seg)      & 0.0 & 0.0 & 0.0 & 0.0 & 0.0 & 0.0 & 0.0 \\
PSALM                  & 0.0 & 0.0 & 0.0 & 0.0 & 0.0 & 0.0 & 0.0 \\
X-Decoder              & 64.3 & 80.0 & 10.5 & 5.6 & 50.0 & 50.0 & 38.7 \\
SEEM                   & 71.4 & 100.0 & 5.3 & 11.1 & 50.0 & 25.0 & 29.2 \\
Florence-2-Large       & 28.6 & 100.0 & 0.0 & 5.6 & 0.0 & 22.9 & 19.8 \\
SAM~3                  & 100.0 & 100.0 & 78.9 & 94.4 & 100.0 & 95.8 & 93.4 \\
Grounded-SAM           & 0.0 & 20.0 & 0.0 & 0.0 & 0.0 & 0.0 & 0.9 \\
OWLv2-Large + SAM      & 85.7 & 100.0 & 68.4 & 72.2 & 100.0 & 56.2 & 67.9 \\
\bottomrule
\end{tabular}
\end{table}

\paragraph{Summary of T2 diagnostics.}
The T2 diagnostics support three observations consistent with the benchmark design. First, source and family breakdowns reveal strong domain dependence: models that perform well on larger or visually salient fruit targets do not necessarily transfer to small disease regions or dense crop/weed scenes. Second, positive-query mask quality and target-absent abstention are often traded off, as the per-query-type and empty-breakdown tables make explicit. Third, rank-based and small-target programs remain particularly difficult because they require both correct referent selection and accurate mask extraction. These findings support reporting T2 with positive-query and target-absent metrics kept separate, augmented by source- and family-level auxiliary diagnostics for auditing overall behavior.

\subsection{Failure Taxonomy and Qualitative Examples}
\label{app:qualitative}

To complement the quantitative diagnostics in Appendices~\ref{app:extended_t1}
and~\ref{app:extended_t2}, we summarize a failure taxonomy distilled from
per-query prediction inspection. The goal is not to introduce additional
metrics, but to identify recurring qualitative patterns that explain the
aggregate behavior of current models on AgroVG. Failures are grouped by the
capability being tested.

\paragraph{Cardinality and set-completeness errors.}
A common T1 failure is predicting the wrong number of boxes. For single-target
queries, models often identify the correct family but return multiple plausible
instances, causing strict accuracy to fail even when one predicted box overlaps
the target. For multi-target queries, models frequently return only the most
salient or largest instances and miss smaller or partially occluded targets.
This behavior is most visible in \texttt{multi\_family\_all} and top-$K$
programs, where the task requires recovering a complete query-conditioned set
rather than detecting a representative object.

\paragraph{Existence-aware abstention errors.}
Inspection of target-absent failures suggests that hallucination is often driven by category cue alone rather than by visual evidence: when the query mentions a familiar agricultural family, many models produce predictions even when no instance is visually present in the scene. This pattern explains the strong inter-model variation in Empty Accuracy reported in Appendices~\ref{app:extended_t1} and~\ref{app:extended_t2}, and supports reporting target-absent behavior separately from positive-query localization.

\paragraph{Rank-binding and spatial selection errors.}
Rank-based templates require models to bind language such as ``leftmost,''
``rightmost,'' ``largest,'' or ``smallest'' to a specific instance. Many models
can detect the correct family but select the wrong ranked instance, especially
in dense scenes with many visually similar plants, fruits, or disease regions.
Side-selection and relation templates show a related failure mode: models often
return all visible instances of the requested family instead of the subset
satisfying the spatial constraint. These errors indicate that open-vocabulary
recognition alone is insufficient for generalized agricultural grounding.

\paragraph{Mask boundary and small-target failures.}
T2 failures often involve approximately correct localization but poor mask
precision. Foreground may be placed near the intended target while leaking into
adjacent leaves, soil, fruit clusters, or neighboring plants; conversely, only
the most salient part of an irregular target may be captured, missing thin,
occluded, or fragmented regions. Small or visually weak targets such as disease
spots, young weeds, and partially occluded fruits compound this problem: in T1
they lead to missed boxes or boxes centered on visually salient but incorrect
instances, and in T2 they lead to fragmented masks, background leakage, or
complete omission. The widening gap between IoU@0.50 and IoU@0.75 in
Appendix~\ref{app:extended_t2} reflects these boundary-precision and
small-target failures, which are especially relevant for downstream use cases
such as early-stage weed control and lesion detection.

\paragraph{Source-conditioned failures.}
Failure patterns vary across sources and target families. Models that work well
on large, color-contrasting fruit targets may not transfer to dense crop/weed
scenes or small plant-disease regions; conversely, models that abstain reliably
on one source may hallucinate foreground on another. This source dependence
motivates the per-source and per-family reporting in Appendices~\ref{app:extended_t1}
and~\ref{app:extended_t2}: aggregate scores alone can hide source-specific
weaknesses that matter for deployment in particular crops, sensors, or field
conditions.

\paragraph{Implications for AgroVG.}
The observed failure modes align with the benchmark's multi-axis design.
Single-target queries test referent binding under minimal cardinality ambiguity.
Multi-target queries expose set-completeness failures invisible in single-object
grounding benchmarks. Target-absent queries reveal hallucination and abstention
calibration. Mask grounding exposes boundary and coverage errors that box
metrics cannot measure. Together, these qualitative patterns justify evaluating
AgroVG along multiple axes (task, query regime, source, family, program type)
rather than reducing agricultural visual grounding to a single aggregate score.

\section{Hosting, Licensing, Maintenance, and Limitations}
\label{app:hosting}

\paragraph{Dataset hosting and access.}
AgroVG is publicly hosted on the Hugging Face Hub at 
\url{https://huggingface.co/datasets/sauryrs/AgroVG}, with the 
preprocessing, evaluation, and benchmarking code released at 
\url{https://anonymous.4open.science/r/AgroVG-5172/} during review and 
to be de-anonymized upon acceptance, alongside an archival snapshot 
with a persistent identifier. The dataset release contains query files, 
dev/test splits, normalized metadata, mask provenance fields, 
evaluation inputs, released prediction files when permitted, and a 
Croissant metadata file describing dataset structure, field-level 
provenance, and split organization.

\paragraph{Licensing.}
AgroVG follows the access and redistribution terms of each source 
dataset: every image and source-derived annotation inherits the 
license and usage conditions of its original dataset 
(Appendix~\ref{app:source_datasets}, Table~\ref{tab:source_datasets}); 
when raw-image redistribution is not permitted or is uncertain, 
AgroVG releases only derived metadata, preprocessing scripts, 
annotation-grounded query records, and evaluation files, and users 
obtain raw images from the original source. AgroVG-specific 
contributions---the unified schema, derived masks, audit tables, 
query templates, split files, and evaluation protocols---are released 
under CC BY 4.0 where compatible with source terms, and the 
benchmarking code is released under the MIT license. A per-source 
license and attribution summary is included in the dataset card.

\paragraph{Maintenance plan.}
We will maintain AgroVG for at least three years after publication, 
including fixing reproducibility issues, releasing patched metadata 
when source-side errata are reported, documenting version history, 
and accepting community-contributed predictions for a public 
leaderboard. Issue tracking, contribution guidelines, and maintainer 
contacts are provided in the dataset card and code repository.

\paragraph{Limitations.}
We summarize the main limitations of AgroVG; additional context 
appears in the cited subsections. \emph{(i) Family coverage of T2:} 
T2 currently instantiates instance-mask grounding on three of the six 
target families (crop/weed, fruit, plant disease) because pest, 
wheat-head, and tree-canopy sources do not yet provide audited 
instance-level pixel evidence under the AgroVG normalization protocol 
(Appendix~\ref{app:source_datasets}). \emph{(ii) Single-reviewer 
expert review:} each candidate image is reviewed once; we mitigate 
this by restricting review criteria to visually evident annotation 
defects and by releasing both retained and dropped image identifiers 
for future audits (Appendix~\ref{app:sampling_qc}). \emph{(iii) 
Template-based query generation:} queries are generated from a 
deterministic, annotation-grounded template library, which gives 
exact target-set traceability but limits linguistic diversity 
(Appendix~\ref{app:query_generation}). \emph{(iv) Geographic and 
sensor coverage:} AgroVG inherits the geographic, sensor, and 
crop-variety distribution of its contributing source datasets and 
should not be treated as evidence of deployment readiness 
(Appendix~\ref{app:broader_impact}). \emph{(v) Single-pass evaluation:} 
results are single-pass zero-shot scores under deterministic or 
provider-default decoding; per-query results, model identifiers, 
and query dates are released to support future re-evaluation. 
\emph{(vi) Scope of evaluated models:} the 26 evaluated 
configurations are representative but not exhaustive, and the public 
leaderboard accepts community-contributed predictions for additional 
configurations.

\section{Case Studies}
\label{app:case_studies}

\begin{figure}[t]
  \centering
  \includegraphics[width=0.98\textwidth]{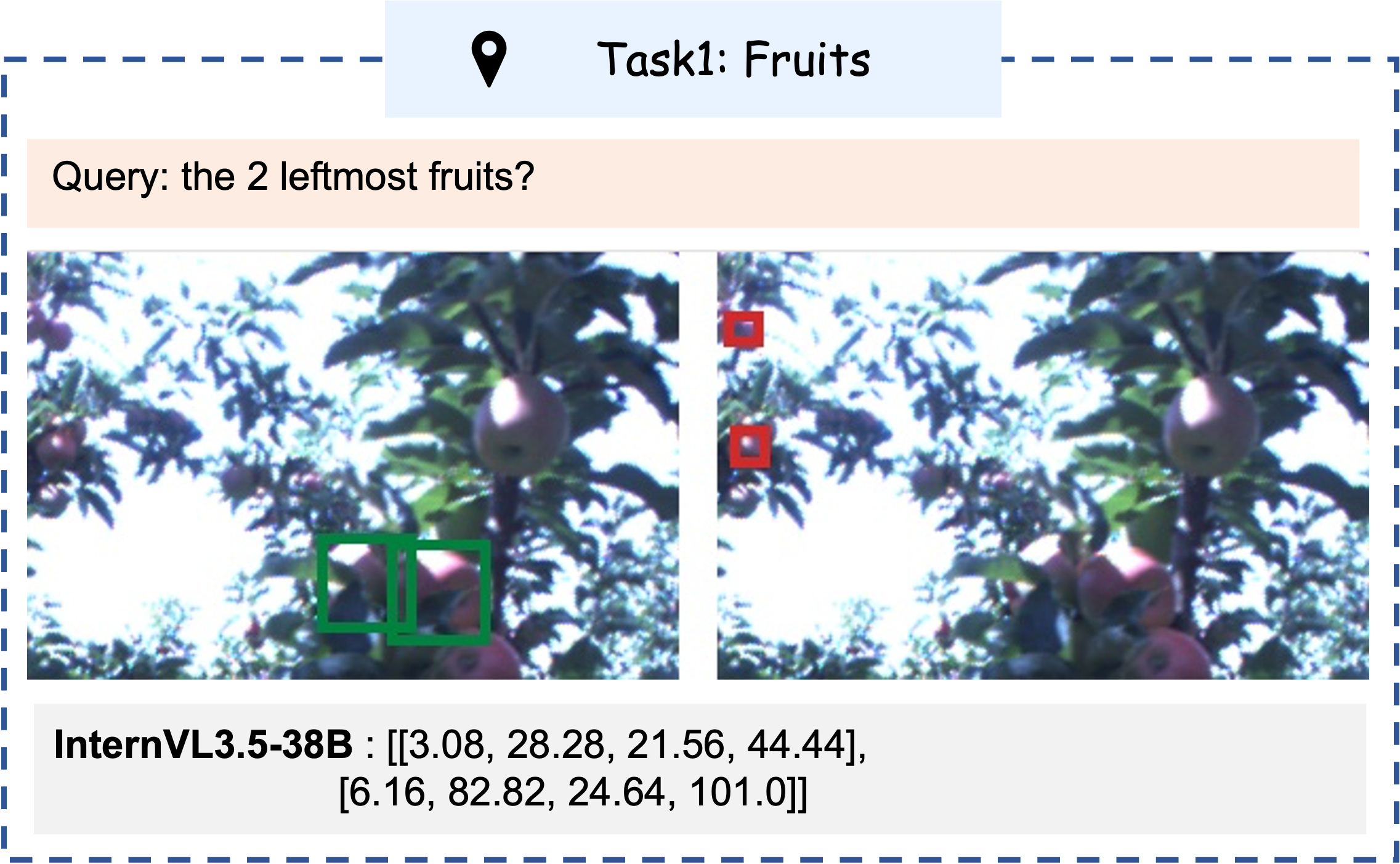}
  \caption{
  \textbf{T1 case study: spatial-rank multi-target query on fruit.}
  Query: ``the two leftmost fruits.'' InternVL3.5-38B fails to bind 
  the spatial-rank constraint to the correct instances, predicting 
  small boxes on background regions rather than the two leftmost 
  fruits in the ground-truth target set. The case illustrates the 
  rank-binding failure mode discussed in 
  Appendix~\ref{app:qualitative}.
  }
  \label{fig:case_study_t1_fruit}
\end{figure}

\begin{figure}[t]
  \centering
  \includegraphics[width=0.98\textwidth]{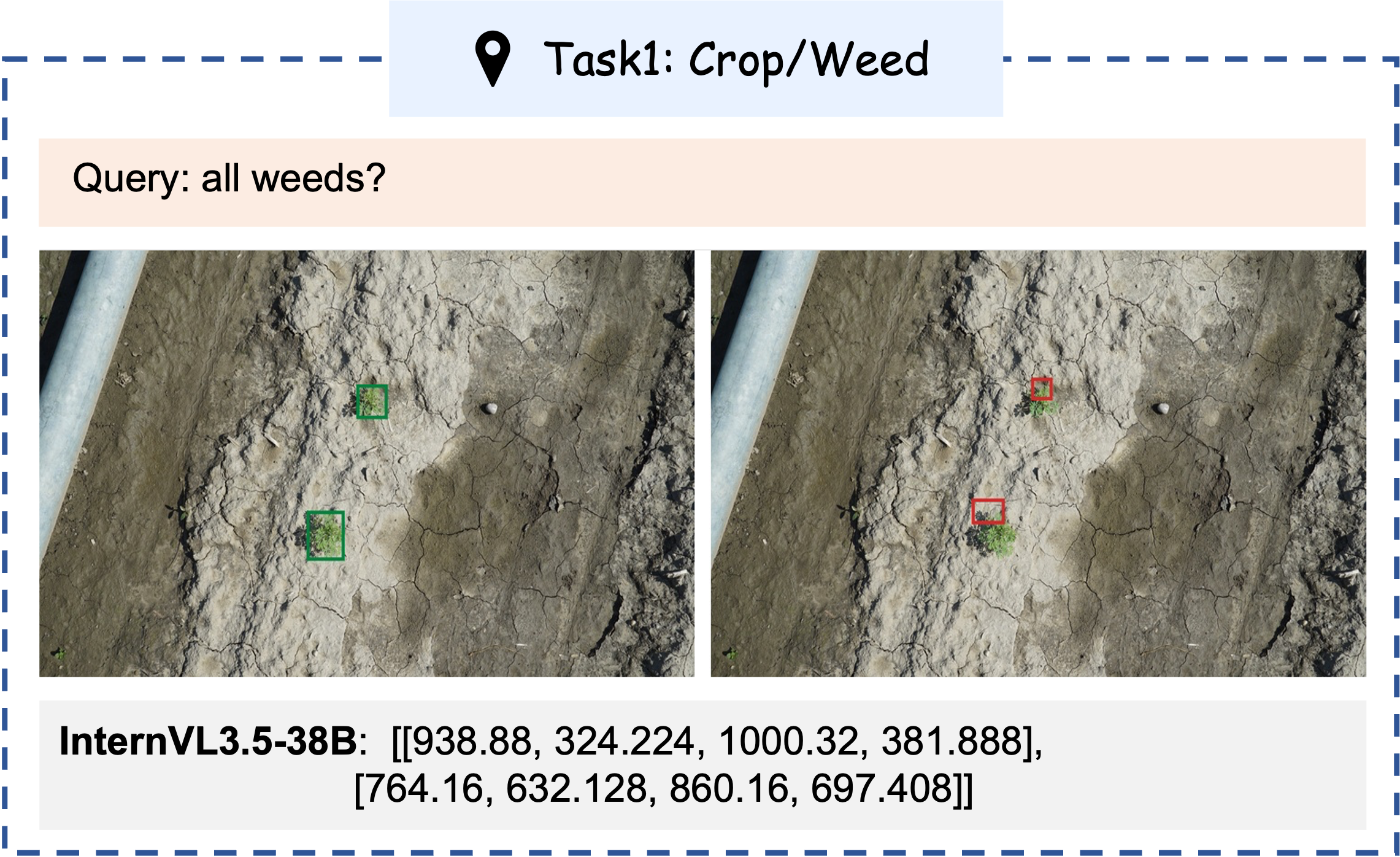}
  \caption{
  \textbf{T1 case study: multi-target query on weed.}
  InternVL3.5-38B detects the correct number of weed instances, but 
  the predicted boxes are shifted and undersized relative to the 
  ground truth, missing the full target extent. The case illustrates 
  that approximate localization can still fail strict IoU matching 
  on densely packed agricultural targets.
  }
  \label{fig:case_study_t1_weed}
\end{figure}

\begin{figure}[t]
  \centering
  \includegraphics[width=0.98\textwidth]{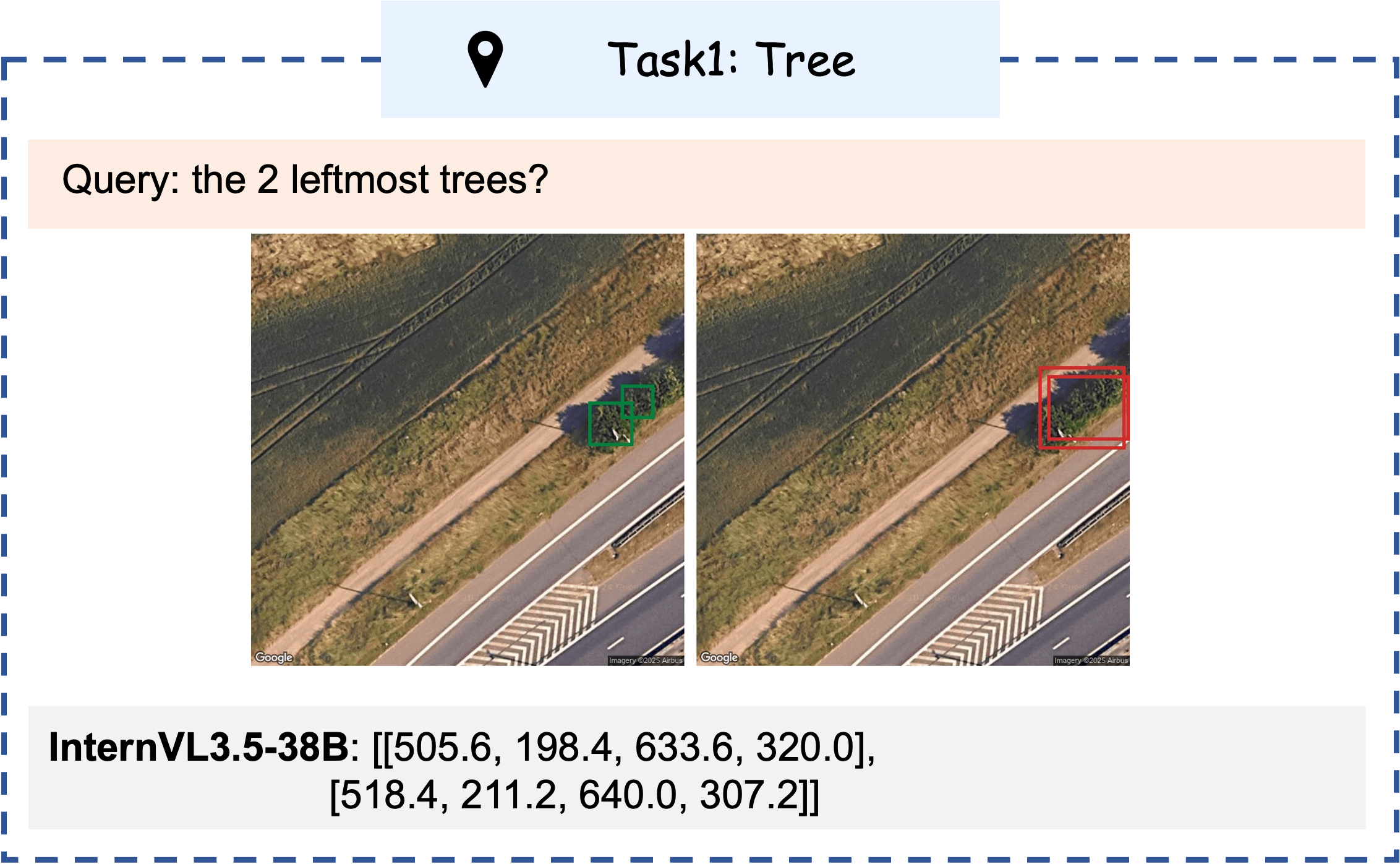}
  \caption{
  \textbf{T1 case study: spatial-rank multi-target query on tree canopy.}
  Query: ``the two leftmost trees.'' InternVL3.5-38B returns large, 
  overlapping boxes that cover the wrong tree region, indicating 
  difficulty with ordinal references in dense canopy scenes where 
  many trees are visually similar. The case illustrates the 
  rank-binding and set-completeness errors discussed in 
  Appendix~\ref{app:qualitative}.
  }
  \label{fig:case_study_t1_tree}
\end{figure}

\begin{figure}[t]
  \centering
  \includegraphics[width=0.98\textwidth]{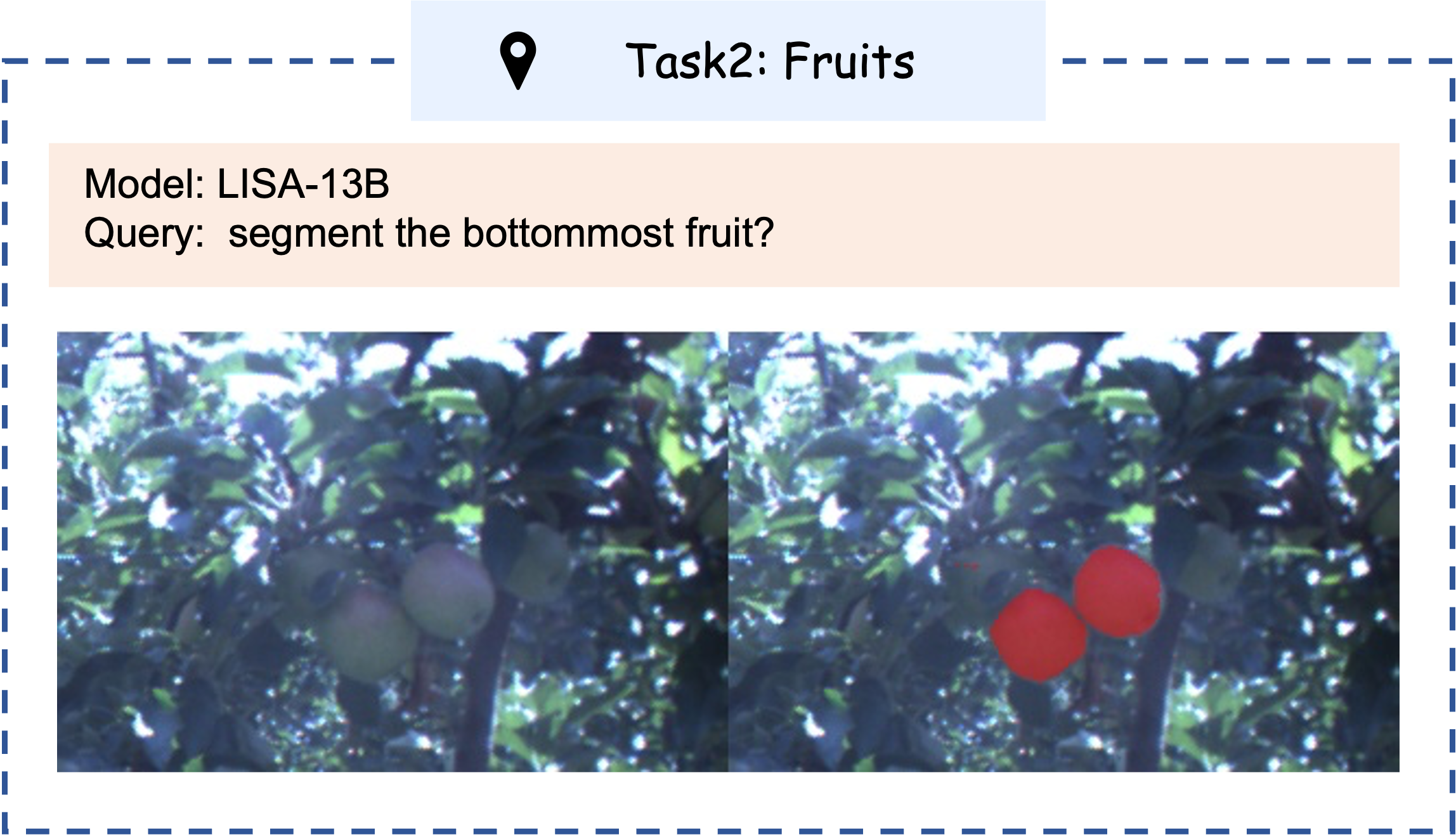}
  \caption{
  \textbf{T2 case study: single-target rank query on fruit.}
  Query: ``the bottommost fruit.'' LISA-13B over-segments the scene 
  by masking multiple fruit regions instead of isolating the unique 
  bottommost instance specified by the referring expression. The case 
  illustrates a cardinality failure on T2: the model produces a 
  reasonable foreground mask but does not honor the single-target 
  constraint encoded in the query.
  }
  \label{fig:case_study_t2_fruit_single}
\end{figure}

\begin{figure}[t]
  \centering
  \includegraphics[width=0.98\textwidth]{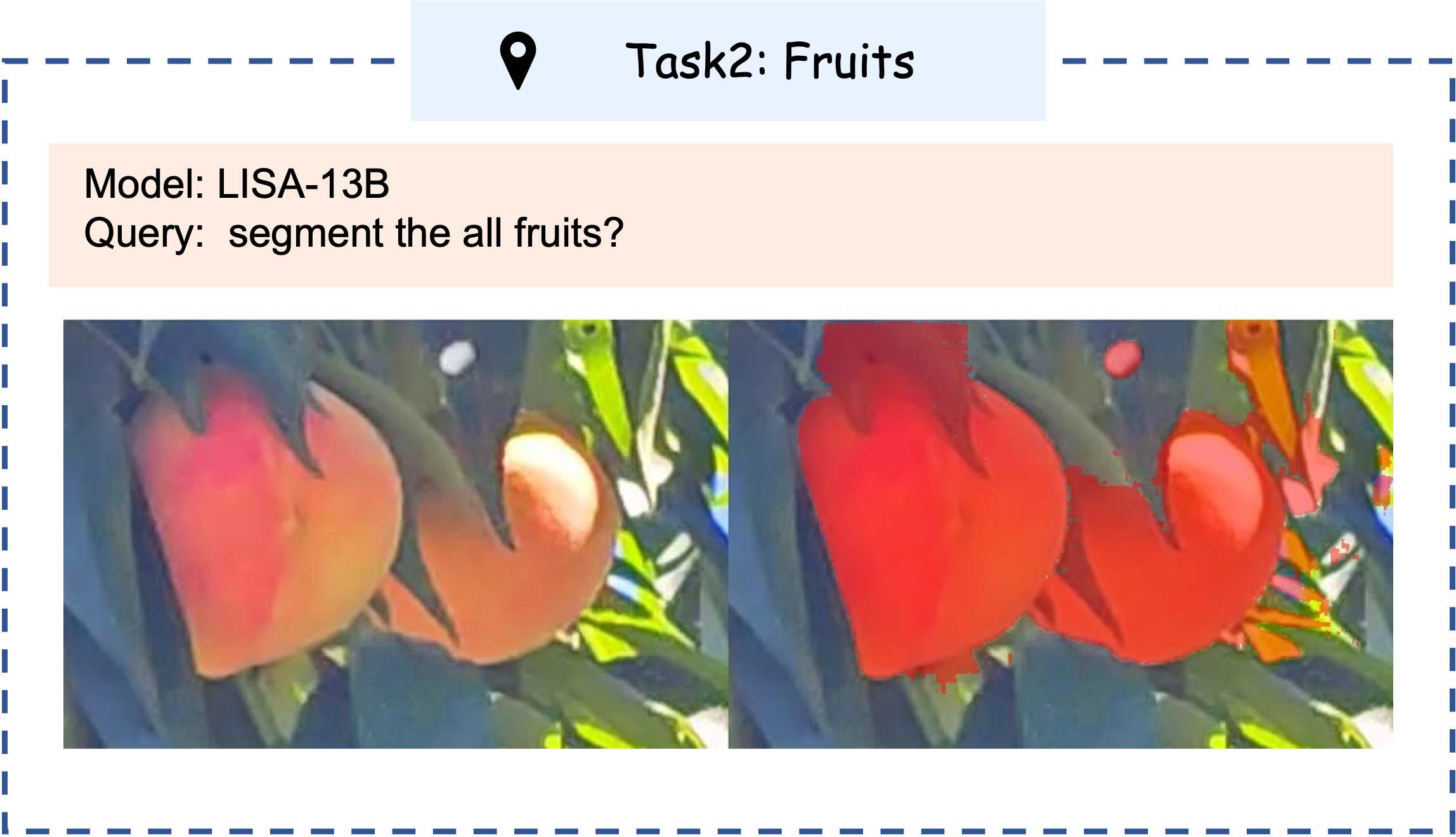}
  \caption{
  \textbf{T2 case study: multi-target query on fruit.}
  LISA-13B identifies the queried fruit regions, but the predicted 
  mask leaks into nearby leaves and background, reducing pixel-level 
  precision. The case illustrates the boundary-leakage failure mode 
  responsible for the gap between IoU@0.50 and IoU@0.75 on T2 reported 
  in Appendix~\ref{app:extended_t2}.
  }
  \label{fig:case_study_t2_fruit_multi}
\end{figure}

\begin{figure}[t]
  \centering
  \includegraphics[width=0.98\textwidth]{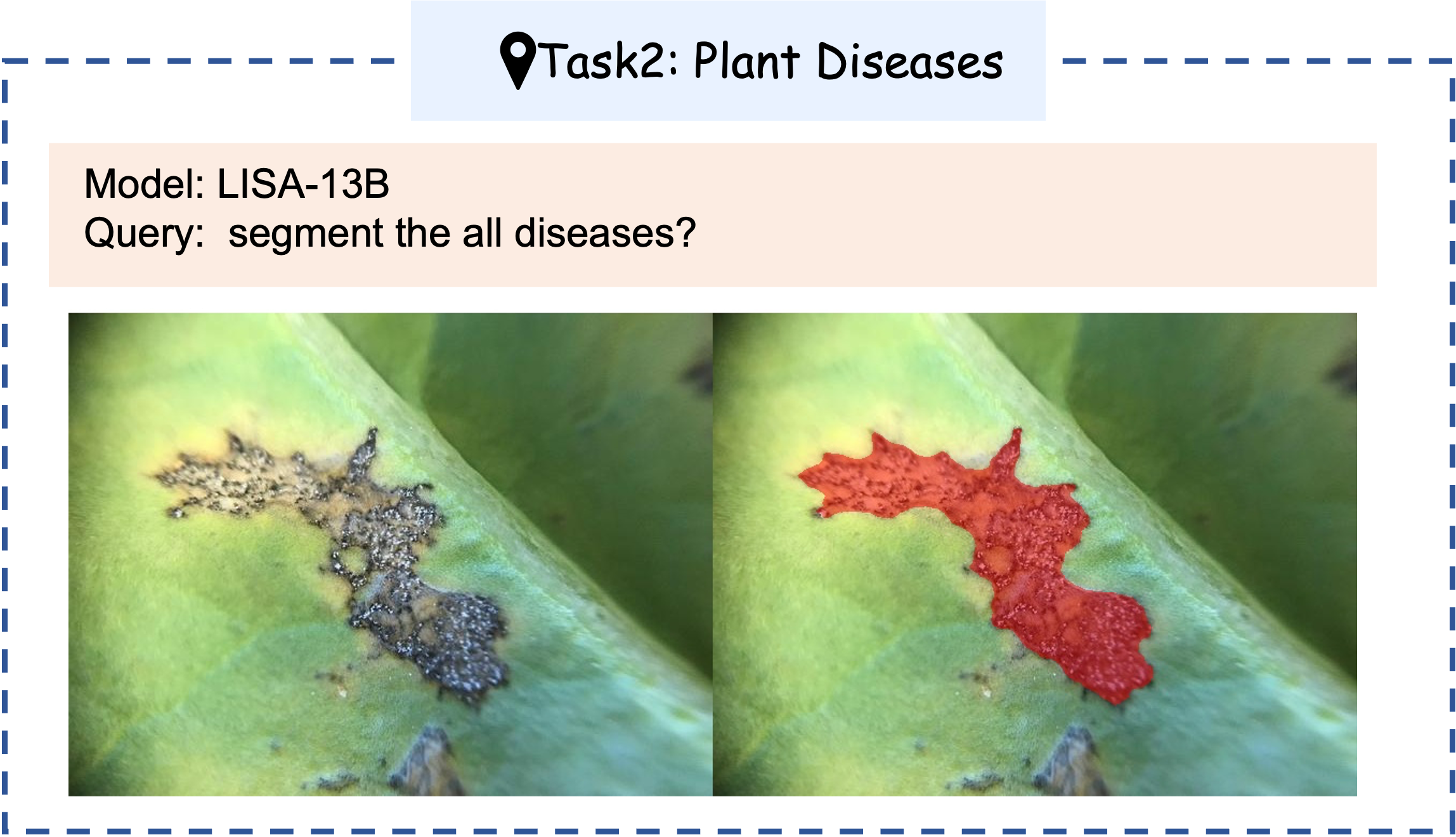}
  \caption{
  \textbf{T2 case study: query on plant disease.}
  LISA-13B's predicted mask roughly covers the diseased area but 
  expands beyond the lesion boundary into healthy leaf tissue. The 
  case illustrates the boundary-precision failures on irregular 
  small targets discussed in Appendix~\ref{app:qualitative}, which 
  are particularly relevant for downstream uses such as early lesion 
  detection.
  }
  \label{fig:case_study_t2_disease}
\end{figure}

In this section, we provide qualitative case studies illustrating how 
representative models behave on AgroVG queries 
(Figs.~\ref{fig:case_study_t1_fruit}--\ref{fig:case_study_t2_disease}). 
We focus on InternVL3.5-38B for T1 bounding-box grounding and LISA-13B 
for T2 instance-mask grounding. InternVL3.5-38B is the strongest model 
on AgroVG-T1 across the evaluated configurations, achieving the best 
overall and family-macro Set-$F_1$ as well as the best strict 
single-target accuracy (Table~\ref{tab:t1_main_results}); it represents 
the current frontier of open-source VLMs on agricultural box grounding. 
LISA-13B is the strongest model on AgroVG-T2 in query-averaged and 
family-macro mIoU (Table~\ref{tab:t2_main_results}); it represents the 
state of the art among segmentation MLLMs and reasoning-segmentation 
models on agricultural mask grounding. By visualizing the predictions 
of these two leading models, we aim to characterize both the upper 
performance bound of current grounding systems on AgroVG and the 
recurring failure modes that remain even at this level.

We present three T1 cases and three T2 cases drawn from the AgroVG 
\texttt{test} split, covering fruit, weed, tree-canopy, and 
plant-disease targets across single-target and multi-target regimes. 
Each case shows an agricultural image, an associated referring 
expression, the ground-truth target set, and the model-generated 
prediction overlaid on the image, allowing direct comparison between 
prediction and ground truth.

Through these examples, we observe that even the strongest evaluated 
models exhibit consistent failure patterns rather than uniformly strong 
performance. On T1, InternVL3.5-38B often identifies the correct target 
family but localizes the wrong instances under spatial-rank queries 
(Figs.~\ref{fig:case_study_t1_fruit} and~\ref{fig:case_study_t1_tree}) 
and produces shifted or undersized boxes that fail to capture the full 
extent of agricultural targets 
(Fig.~\ref{fig:case_study_t1_weed}). On T2, LISA-13B produces 
plausible mask shapes but suffers from poor boundary precision: it 
over-segments multiple instances when only one is queried 
(Fig.~\ref{fig:case_study_t2_fruit_single}), leaks foreground into 
adjacent leaves and background on multi-fruit scenes 
(Fig.~\ref{fig:case_study_t2_fruit_multi}), and expands disease masks 
into healthy leaf tissue 
(Fig.~\ref{fig:case_study_t2_disease}). These observations are 
consistent with the quantitative diagnostics reported in 
Appendices~\ref{app:extended_t1} and~\ref{app:extended_t2} and the 
failure taxonomy in Appendix~\ref{app:qualitative}, and they show that 
strong aggregate scores on AgroVG do not imply uniform competence 
across query regimes, target families, or instance densities.

\clearpage
\input{checklist.tex}

\end{document}

%% file: checklist.tex
\section*{NeurIPS Paper Checklist}

\begin{enumerate}

\item {\bf Claims}
    \item[] Question: Do the main claims made in the abstract and introduction accurately reflect the paper's contributions and scope?
    \item[] Answer: \answerYes{}
    \item[] Justification: The abstract and introduction state the benchmark scope, dataset size, source coverage, T1/T2 protocols, query regimes, and zero-shot evaluation setting. The claims are supported by the benchmark construction, evaluation protocol, main results, and stated limitations in the paper and appendix.

\item {\bf Limitations}
    \item[] Question: Does the paper discuss the limitations of the work performed by the authors?
    \item[] Answer: \answerYes{}
    \item[] Justification: The conclusion and Appendix~\ref{app:query_generation} discuss dataset-construction limitations, including restricted T2 family coverage, single-reviewer expert screening, and template-based query generation. The paper also discusses deployment-relevant failure modes in Appendix~\ref{app:qualitative}.

\item {\bf Theory assumptions and proofs}
    \item[] Question: For each theoretical result, does the paper provide the full set of assumptions and a complete (and correct) proof?
    \item[] Answer: \answerNA{}
    \item[] Justification: The paper introduces a benchmark, data construction pipeline, and evaluation protocol, and does not present theoretical results, theorems, or proofs.

    \item {\bf Experimental result reproducibility}
    \item[] Question: Does the paper fully disclose all the information needed to reproduce the main experimental results of the paper to the extent that it affects the main claims and/or conclusions of the paper (regardless of whether the code and data are provided or not)?
    \item[] Answer: \answerYes{}
    \item[] Justification: Appendix~\ref{app:evaluation_setup} specifies metric definitions, model interfaces, prompt templates, parsing rules, compute environment, and released artifacts. Appendix~\ref{app:compute} describes score recomputation from released predictions and full inference reproduction using the documented protocols.

\item {\bf Open access to data and code}
    \item[] Question: Does the paper provide open access to the data and code, with sufficient instructions to faithfully reproduce the main experimental results, as described in supplemental material?
    \item[] Answer: \answerYes{}
    \item[] Justification: Appendix~\ref{app:hosting} provides public dataset and code locations, and Appendix~\ref{app:compute} describes the released data, protocol files, predictions, and evaluator outputs. Raw images remain governed by source-dataset licenses where redistribution is restricted.

\item {\bf Experimental setting/details}
    \item[] Question: Does the paper specify all the training and test details (e.g., data splits, hyperparameters, how they were chosen, type of optimizer) necessary to understand the results?
    \item[] Answer: \answerYes{}
    \item[] Justification: The experiments are zero-shot evaluations without training or fine-tuning; the paper specifies dev/test split construction, model configurations, prompt and parsing protocols, threshold selection on the development split, decoding settings, and hardware details in the main experimental setup and Appendix~\ref{app:evaluation_setup}.

\item {\bf Experiment statistical significance}
    \item[] Question: Does the paper report error bars suitably and correctly defined or other appropriate information about the statistical significance of the experiments?
    \item[] Answer: \answerNo{}
    \item[] Justification: The paper does not report error bars, confidence intervals, or statistical significance tests because the results are single-pass zero-shot evaluations on a fixed expert-reviewed test split. Instead, it releases per-query results and reports source-, family-, query-type-, and program-level breakdowns for diagnostic auditing.

\item {\bf Experiments compute resources}
    \item[] Question: For each experiment, does the paper provide sufficient information on the computer resources (type of compute workers, memory, time of execution) needed to reproduce the experiments?
    \item[] Answer: \answerYes{}
    \item[] Justification: Appendix~\ref{app:compute} reports that local open-source and specialized configurations were evaluated on NVIDIA A100 80GB GPUs and that the local evaluation campaign required approximately 320 A100 GPU-hours. It also states that closed-source API compute is provider-side and not observable from the client.
    
\item {\bf Code of ethics}
    \item[] Question: Does the research conducted in the paper conform, in every respect, with the NeurIPS Code of Ethics \url{https://neurips.cc/public/EthicsGuidelines}?
    \item[] Answer: \answerYes{}
    \item[] Justification: The work uses public agricultural datasets, releases derived benchmark artifacts under documented license constraints, and contains no human subjects, facial imagery, or personally identifiable information. Potential risks and responsible-use considerations are discussed in Appendix~\ref{app:broader_impact}.

\item {\bf Broader impacts}
    \item[] Question: Does the paper discuss both potential positive societal impacts and negative societal impacts of the work performed?
    \item[] Answer: \answerYes{}
    \item[] Justification: Appendix~\ref{app:broader_impact} discusses positive impacts for precision agriculture and agricultural monitoring, as well as risks from incorrect interventions, transfer to surveillance-like monitoring, and labor-market effects. It also recommends disaggregated reporting and validation before deployment.
    
\item {\bf Safeguards}
    \item[] Question: Does the paper describe safeguards that have been put in place for responsible release of data or models that have a high risk for misuse (e.g., pre-trained language models, image generators, or scraped datasets)?
    \item[] Answer: \answerNA{}
    \item[] Justification: The paper releases a benchmark dataset and evaluation code, not a high-risk pretrained model, image generator, or scraped dataset of people. Responsible-use guidance and source-license constraints are nevertheless discussed in Appendices~\ref{app:broader_impact} and~\ref{app:hosting}.

\item {\bf Licenses for existing assets}
    \item[] Question: Are the creators or original owners of assets (e.g., code, data, models), used in the paper, properly credited and are the license and terms of use explicitly mentioned and properly respected?
    \item[] Answer: \answerYes{}
    \item[] Justification: Source datasets and evaluated models are cited in the paper, and Appendix~\ref{app:source_datasets} and Appendix~\ref{app:hosting} describe source provenance, access constraints, and licensing. The released dataset card includes a per-source license and attribution summary.

\item {\bf New assets}
    \item[] Question: Are new assets introduced in the paper well documented and is the documentation provided alongside the assets?
    \item[] Answer: \answerYes{}
    \item[] Justification: AgroVG introduces new benchmark assets, including unified metadata, query records, split files, derived masks where applicable, evaluation inputs, prompts, parsers, and benchmarking code. Their schema, construction pipeline, licenses, hosting, and maintenance plan are documented in Appendices~\ref{app:dataset_construction},~\ref{app:evaluation_setup}, and~\ref{app:hosting}.

\item {\bf Crowdsourcing and research with human subjects}
    \item[] Question: For crowdsourcing experiments and research with human subjects, does the paper include the full text of instructions given to participants and screenshots, if applicable, as well as details about compensation (if any)? 
    \item[] Answer: \answerNA{}
    \item[] Justification: The paper does not involve crowdsourcing experiments or human-subject research. Expert visual review is used only for agricultural image and annotation quality control, as described in Appendix~\ref{app:sampling_qc}.

\item {\bf Institutional review board (IRB) approvals or equivalent for research with human subjects}
    \item[] Question: Does the paper describe potential risks incurred by study participants, whether such risks were disclosed to the subjects, and whether Institutional Review Board (IRB) approvals (or an equivalent approval/review based on the requirements of your country or institution) were obtained?
    \item[] Answer: \answerNA{}
    \item[] Justification: The work does not involve human-subject research, study participants, or personally identifiable information. AgroVG is constructed from public agricultural imagery and source annotations.

\item {\bf Declaration of LLM usage}
    \item[] Question: Does the paper describe the usage of LLMs if it is an important, original, or non-standard component of the core methods in this research? Note that if the LLM is used only for writing, editing, or formatting purposes and does \emph{not} impact the core methodology, scientific rigor, or originality of the research, declaration is not required.
    \item[] Answer: \answerYes{}
    \item[] Justification: The core experimental evaluation includes closed-source and open-source MLLMs/VLMs, and their model identifiers, prompting protocols, parsing rules, and run metadata are documented in the main experiments and Appendix~\ref{app:models_prompts}. Query generation is template-based and annotation-grounded rather than LLM-generated.

\end{enumerate}